\documentclass[twocolumn,aps,pre,superscriptaddress]{revtex4-2}
\usepackage{psfrag,epsfig,amsfonts,amssymb,amsmath,wasysym,bm}
\usepackage{dcolumn}
\usepackage{bbold}
\usepackage[normalem]{ulem}
\usepackage{color}
\usepackage{tabularx, tabulary}
\usepackage{longtable}
\usepackage{tabu}

\usepackage{lipsum}

\usepackage{enumitem} 

\usepackage{hyperref}

\newcolumntype{Z}{>{\raggedleft\arraybackslash}X}

\newcommand{\ket}[1]{\lvert #1 \rangle} 	
\newcommand{\mc}{\mathcal}	

\newcommand{\<}{\langle}	
\renewcommand{\>}{\rangle}	

\newcommand{\given}{\,|\,}

\newcommand{\e}{\mathrm{e}}		
\newcommand{\I}{\mathrm{i}}		

\newcommand{\lmat}{\left( \begin{matrix}}	
\newcommand{\rmat}{\end{matrix} \right)}	

\newcommand{\id}{\mathbb 1}
\newcommand{\RR}{{\mathbb R}}

\newcommand{\ZZ}{{\mathbb Z}}

\newcommand{\mref}[1]{\ref{#1}}
\newcommand{\meqref}[1]{\eqref{#1}}

\newcommand{\NV}{M}
\newcommand{\NH}{N}
\newcommand{\DeltaEmp}{\tilde{\Delta}_\theta}
\newcommand{\DKL}{D_{\mathrm{KL}}}
\newcommand{\Ctot}{C_{\mathrm{tot}}}
\newcommand{\pEmp}{\tilde{p}}
\newcommand{\nCD}{n_{\mathrm{CD}}}

\newcommand{\KPH}{{\mc K}_{\mathrm{H}}}

\begin{document}

\title{Three Learning Stages and Accuracy-Efficiency Tradeoff\\of Restricted Boltzmann Machines}

\author{Lennart Dabelow}
\affiliation{RIKEN Center for Emergent Matter Science (CEMS), Wako, Saitama 351-0198, Japan}

\author{Masahito Ueda}
\affiliation{Department of Physics and Institute for Physics of Intelligence, Graduate School of Science, The University of Tokyo, Bunkyo-ku, Tokyo 113-0033, Japan}
\affiliation{RIKEN Center for Emergent Matter Science (CEMS), Wako, Saitama 351-0198, Japan}

\date{\today}

\begin{abstract}
Restricted Boltzmann Machines (RBMs) offer a versatile architecture for unsupervised machine learning
that can in principle approximate any target probability distribution with arbitrary accuracy.
However, the RBM model is usually not directly accessible due to its computational complexity, 
and Markov-chain sampling is invoked to analyze
the
learned probability
distribution.
For training and eventual applications,
it is thus desirable to have a sampler that is both accurate and efficient.
We highlight that these two goals generally compete with each other and cannot be achieved simultaneously.
More specifically, we identify and quantitatively characterize three regimes of RBM learning:
independent learning, where the accuracy improves
without losing
efficiency;
correlation learning, where higher accuracy entails lower efficiency;
and degradation, where both accuracy and efficiency no longer improve or even deteriorate.
These findings are based on numerical experiments and heuristic arguments.
\end{abstract}

\maketitle


Restricted Boltzmann Machines (RBMs) \cite{Ackley:1985lab, Smolensky:1986ipd} are a
versatile
and conceptionally simple unsupervised machine learning model.
Besides traditional applications,
such as dimensional reduction and pretraining \cite{Hinton:2006rdd, Gehler:2006rap, Hinton:2007rsf, Salakhutdinov:2007rbm} and text classification \cite{Larochelle:2008cud},
they have become increasingly widespread in the physics community \cite{Carleo:2019mlp, Mehta:2019hbl}.
Examples include
tomography \cite{Torlai:2018nnq, Torlai:2018lsp} and variational encoding \cite{Carleo:2017sqm, Nomura:2017rbm, Gao:2017erq, Glasser:2018nnq, Xia:2018qml, Melko:2019rbm, Choo:2020fnn} of quantum states,
time-series forecasting \cite{Kuremoto:2014tsf},
and information-based renormalization group transformations \cite{KochJanusz:2018min, Lenggenhager:2020org}.

\begin{figure}
\includegraphics[scale=1]{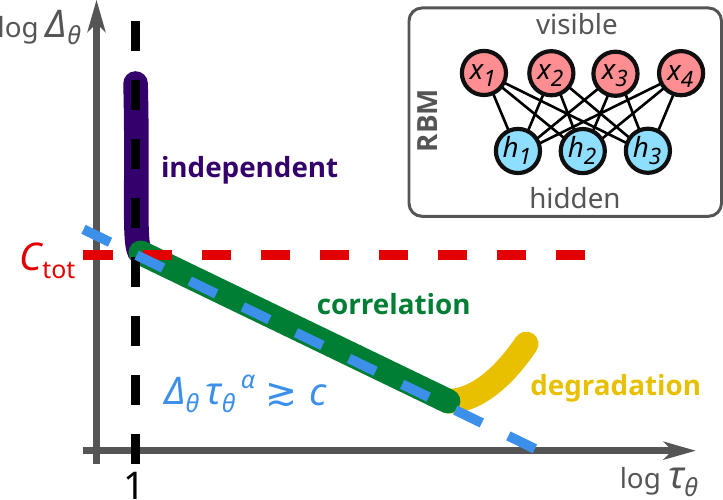}
\caption{Schematic illustration of the three learning regimes of Restricted Boltzmann Machines (RBMs),
characterized by the relationship between the model's divergence $\Delta_\theta$ from the target distribution (accuracy, cf.\ Eq.~\eqref{eq:LossDKL})
and its integrated autocorrelation time $\tau_\theta$ (efficiency, cf.\ Eq.~\eqref{eq:iact}):
independent learning with improved accuracy at no efficiency loss,
correlation learing with a power-law tradeoff relation between accuracy and efficiency,
and the degragation regime with steady or diminishing accuracy and loss of efficiency.
Inset: Schematic illustration of the RBM structure comprised of visible and hidden units.}
\label{fig:Sketch3Regimes}
\end{figure}

A general goal in unsupervised machine learning is to find the best representation of some unknown \emph{target probability distribution} $p(x)$ within a family of \emph{model distributions} $\hat p_\theta(x)$,
where $\theta$ denotes the model parameters to be optimized.
To this end, the RBM architecture introduces two types of \emph{units},
the \emph{visible} units $x = (x_1, \ldots, x_{\NV}) \in \mathcal{X}$,
which relate to the states of the target distribution,
and the \emph{hidden} units $h = (h_1, \ldots, h_{\NH}) \in \mathcal{H}$,
which mediate correlations between the visible units
(see, e.g., Refs.~\cite{Hinton:2012pgt, Fischer:2012irb, Montufar:2018rbm} for reviews
and the top-right corner of Fig.~\ref{fig:Sketch3Regimes} for an illustration).
We focus on the most common case where both the visible and the hidden units are binary,
i.e., $\mathcal{X} = \{ 0, 1 \}^{\NV}$ and $\mathcal{H} = \{ 0, 1 \}^{\NH}$. 
The RBM model is based on a joint Boltzmann distribution for $x$ and $h$,
\begin{equation}
\label{eq:RBM:pHX}
	\hat p_\theta(x, h)
		:= Z_\theta^{-1} \, \e^{ -E_\theta(x, h) } \,,
\end{equation}
where the ``energy'' $E_\theta(x, h) := -\sum_{i,j} w_{ij} x_i h_j - \sum_i a_i x_i - \sum_j b_j h_j$ takes the form of a
classical spin Hamiltonian with ``interactions'' between visible and hidden units described by the \emph{weights} $w_{ij} \in \RR$ and ``external fields'' for
visible and hidden
units
described
by the \emph{biases} $a_i, b_j \in \RR$.
The weights and biases constitute the model parameters $\theta = (w_{ij}, a_i, b_j)$,
and the normalization factor
\begin{equation}
\label{eq:RBM:Z}
	Z_\theta := \sum_{x, h} \e^{ -E_\theta(x, h) }
\end{equation}
is referred to as the \emph{partition function}.
The model distribution $\hat p_\theta(x)$
that approximates
the target $p(x)$ is
obtained from marginalization over the hidden units, $\hat p_\theta(x) := \sum_h \hat p_\theta(x, h)$.

The major drawback of
RBMs
is that the computational cost to evaluate $Z_\theta$ (and hence
$\hat p_\theta(x, h)$ and $\hat p_\theta(x)$) scales exponentially with $\min\{ \NV, \NH \}$ 
(see also Methods),
which renders the model intractable in practice
\cite{Long:2010rbm}.
Therefore,
both training (i.e., finding the optimal $\theta$) and
deployment (i.e.,
applying
a trained model)
rely on
approximate sampling from $\hat p_\theta(x)$,
typically
via Markov chains.
Ideally, one
wishes
to
generate samples
both \emph{efficiently}, in the sense of minimal correlation and computational cost,
and \emph{accurately} in the sense of a faithful representation of the target $p(x)$.
Unfortunately,
these two goals generally compete
and cannot be achieved simultaneously.

In this work,
we
explore the tradeoff relationship between accuracy and efficiency
by identifying three distinct regimes of RBM training
as illustrated in Fig.~\ref{fig:Sketch3Regimes}:
(i)~independent learning, where the accuracy can be improved without 
sacrificing
efficiency;
(ii)~correlation learning, where higher accuracy entails lower efficiency, typically in the form of a power-law tradeoff;
and (iii)~degradation, where limited expressivity, overfitting, and/or approximations in the learning algorithm lead to 
reduced efficiency with no gain or even loss of accuracy.

Biased or inefficient sampling is a known limitation of standard training algorithms \cite{Desjardins:2010tmc, Fischer:2012irb, Decelle:2021ene},
but it is not an artifact of deficient training methods.
Rather, it should be understood as an intrinsic limitation of the RBM model.
Yet its consequences for the usefulness of trained models in applications have received relatively little attention thus far.
Our observations~(i)--(iii) above elucidate the inner workings of RBMs and imply that, depending on the intended applications, aiming at maximal accuracy may not always be beneficial.
We demonstrate the various aspects of these findings
by way of several
problems,
ranging from quantum-state tomography for the transverse-field Ising chain (TFIC, cf.\ Fig.~\ref{fig:Tradeoff:TFIM}) to pattern recognition and image generation (Figs.~\ref{fig:Tradeoff:PH1} and~\ref{fig:Tradeoff:Applications});
see also the figure captions and Methods for more details on the examples.

\section*{Results}

\subsection*{Accuracy and efficiency}

A natural measure for the accuracy of the model distribution $\hat p_\theta(x)$ is its Kullback-Leibler (KL) divergence $\DKL(p||\hat p_\theta)$ \cite{CoverThomas:ElementsOfInformationTheory} with respect to the target distribution $p(x)$,
\begin{equation}
\label{eq:LossDKL}
	\Delta_\theta := \DKL(p||\hat p_\theta) \equiv \sum_x p(x) \, \log \frac{ p(x) }{ \hat p_\theta(x) } \,,
\end{equation}
which is nonnegative and vanishes if and only if the distributions $p(x)$ and $\hat p_\theta(x)$ agree.
Indeed, $\Delta_\theta$ provides the basis of most standard training algorithms for RBMs such as contrastive divergence (CD) \cite{Hinton:2002tpe, Hinton:2012pgt}, persistent CD (PCD) \cite{Tieleman:2008trb}, fast PCD \cite{Tieleman:2009ufw}, or parallel tempering \cite{Salakhutdinov:2009lmr, Desjardins:2010tmc}.
Adopting a gradient-descent scheme with $\Delta_\theta$ as the loss function,
one would ideally update the parameters
according to
\begin{align}
\label{eq:RBM:TrainingUpdate}
	\theta_k(t+1) - \theta_k(t)
		&= - \eta \left[ \left\< \tfrac{\partial E_\theta(x, h)}{\partial \theta_k} \right\>_{\! \hat p_\theta(h \!\given\! x) p(x)} 
		\right. \notag \\
		& \qquad\qquad \left. - \left\< \tfrac{\partial E_\theta(x, h)}{\partial \theta_k} \right\>_{\! \hat p_\theta(x, h)} \right] ,
\end{align}
where $\eta > 0$ is the learning rate and $\hat p_\theta(h \given x)$ is the conditional distribution of the hidden units given the visible ones.
Since this conditional distribution factorizes and the
dependence on $Z_\theta$ cancels out
(see Methods for explicit expressions),
the first average on the right-hand side of~\eqref{eq:RBM:TrainingUpdate} can readily be evaluated.
More precisely, since $p(x)$ is unknown, it needs to be approximated by the empirical distribution $\pEmp(x; S) := \frac{1}{\lvert S \rvert} \sum_{\tilde x \in S} \delta_{x, \tilde x}$ for a (multi)set of \emph{training data} $S := \{ \tilde x^{(1)}, \ldots, \tilde x^{(\lvert S \rvert)} \}$,
which are assumed to be independent samples drawn from
$p(x)$.
Hence the effective loss function is $\DeltaEmp^{(S)} := \sum_x \pEmp(x; S) \log \frac{\pEmp(x; S)}{\hat p_\theta(x)}$,
which is an empirical counterpart of~\eqref{eq:LossDKL}.

The second average in~\eqref{eq:RBM:TrainingUpdate}, however, requires the full model distribution~\eqref{eq:RBM:pHX} and is thus not directly accessible in practice.
Instead,
it is
usually approximated by sampling alternatingly from the accessible conditional distributions $\hat p_\theta(h \given x)$ and $\hat p_\theta(x \given h)$,
leading to a Markov chain of the form
\begin{equation}
\label{eq:MarkovChainXH}
	x^{(0)} \rightarrow h^{(0)} \rightarrow x^{(1)} \rightarrow h^{(1)} \rightarrow \cdots
\end{equation}
The distribution of $(x^{(n)}, h^{(n)})$ converges to the model distribution $\hat p_\theta(x, h)$ as $n \to \infty$.
Truncating the chain~\eqref{eq:MarkovChainXH} at a finite $n = \nCD$,
we obtain
a (biased) sample from that distribution,
whose bias vanishes as $\nCD \to \infty$ \cite{Bengio:2009jgc},
but depends on the initialization of the chain for finite $\nCD$.
In our numerical examples, we will usually adopt the common CD algorithm,
which chooses $x^{(0)}$ as a sample from the training data $S$,
or the PCD algorithm,
where $x^{(0)}$ is a sample from the chain of the previous update step
(see also Supplementary Note~1).
Subsequently,
the Markov chain~\eqref{eq:MarkovChainXH} can be used to generate a new, 
but correlated sample.
Similarly, when analyzing and deploying a model $\hat p_\theta(x)$ after training,
new samples
are typically generated
by means of Markov chains~\eqref{eq:MarkovChainXH},
with the caveat that those samples are correlated and thus not independent.

To quantify the \emph{sampling efficiency}, we therefore consider the integrated autocorrelation time \cite{Sokal:1997mcm}
\begin{equation}
\label{eq:iact}
	\tau_\theta := 1 + 2 \sum_{n=1}^\infty \frac{g_\theta(n)}{g_\theta(0)} \,,
\end{equation}
where $g_\theta(n) := \frac{1}{\NV} \sum_i [ \< x^{(0)}_i x^{(n)}_i \> - \< x^{(0)}_i \>^2 ]$ is the mean correlation function of the visible units for the Markov chain~\eqref{eq:MarkovChainXH} in the stationary regime,
i.e., $x^{(0)} \sim \hat p_\theta(x)$.
Notably, $\tau_\theta$ is independent of the training algorithm
since
it depends only on the RBM parameters $\theta$,
but not on the different initialization schemes of the Markov chains in (P)CD and its variants.
In practice, particularly when utilizing the scheme~\eqref{eq:MarkovChainXH} to employ a trained model productively,
one will start from an arbitrary distribution and discard a number of initial samples (ideally on the order of the mixing time \cite{Bengio:2009jgc, Fischer:2015bcr, Tosh:2016mra}) to thermalize the chain and approach the stationary distribution $\hat p_\theta(x)$.

The interpretation of $\tau_\theta$ as a measure of sampling efficiency is as follows:
Suppose we have a number $R$ of \emph{independent} samples from the model distribution $\hat p_\theta(x)$ to estimate $\< x_i \>$ (or $\frac{1}{M} \sum_i \< x_i \>$).
To obtain an estimate of the same quality via Gibbs sampling according to~\eqref{eq:MarkovChainXH},
we would then need
on the order
of $\tau_\theta R$ correlated Markov-chain samples (see, for example, Sec.~2 of Ref.~\cite{Sokal:1997mcm} and also Methods).
Hence the (minimal) value of $\tau_\theta = 1$ hints at independent (uncorrelated) samples,
and the larger $\tau_\theta$ becomes,
the more samples are needed in principle, rendering the approach less efficient.

Note that the integrated autocorrelation time $\tau_\theta$ defined in Eq.~\eqref{eq:iact} is conceptually related to, but different from the mixing time of the Markov chain
(see also Discussion below).
Furthermore,
different observables (i.e., functions of the visible units $x_i$) generally exhibit different autocorrelation times.
As explained in
detail in Methods,
the quantity $\tau_\theta$ from~\eqref{eq:iact} is a weighted average of the autocorrelation times associated with the observables' elementary variables, namely the individual $x_i$.
Hence we expect $\tau_\theta$ to capture the relevant correlations and thus the sampling efficiency in the generic case.
The evaluation of other correlation measures introduced below will reinforce this notion.
In addition, a quantitative comparison of autocorrelation times for different observables is provided in Supplementary Note~4 for the examples from Figs.~\ref{fig:Tradeoff:TFIM} and~\ref{fig:Tradeoff:Applications}a--c.

Our principal object of study is the mutual dependence of $\Delta_\theta$ and $\tau_\theta$ on the parameters $\theta$
for a given target distribution $p(x)$.
As outlined above and illustrated in Fig.~\ref{fig:Sketch3Regimes},
there are three regimes the machine undergoes during the learning process.
Globally, the overall tradeoff between accuracy and efficiency is
numerically
found
to be bounded by a power law of the form
\begin{equation}
\label{eq:Tradeoff}
	\Delta_\theta \, \tau_\theta^{\;\alpha} \gtrsim c\,,
\end{equation}
where both $c$ and the exponent $\alpha$ are positive constants whose meaning will be clarified in the following.
Moreover,
in the correlation-learning regime,
$\Delta_\theta$ and $\tau_\theta$ are often directly related by a power law $\Delta_\theta \tau_\theta^{\,\alpha'} \simeq c'$,
where the constants $c'$ and $\alpha'$ are close to $c$ and $\alpha$, respectively.

\begin{figure*}
\centering
\includegraphics[scale=1]{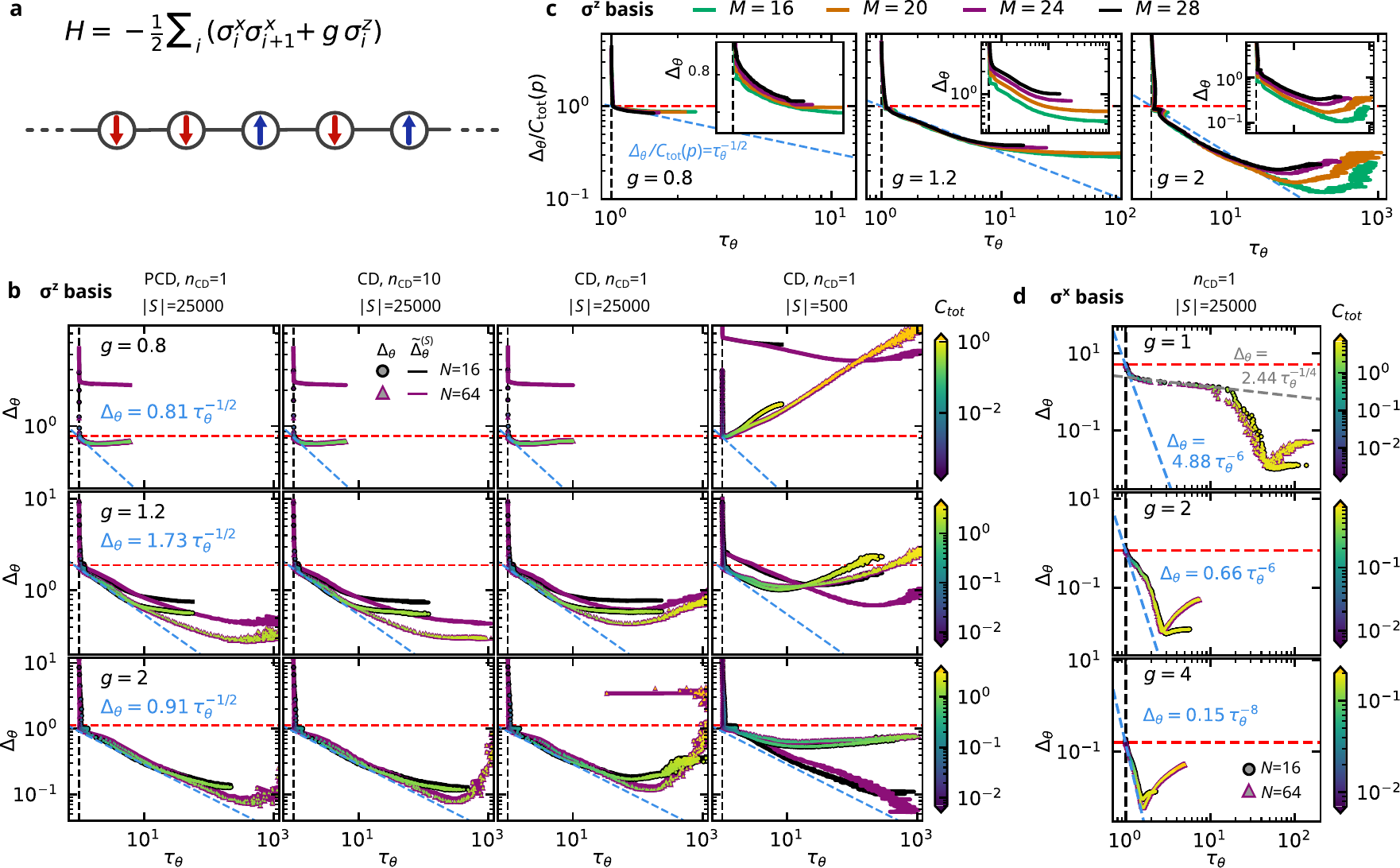}
\caption{RBM learning characteristics for a quantum state tomography task.
The ground state of the transverse-field Ising chain with $\NV$ lattice sites is reconstructed from magnetization measurements along a fixed axis,
namely the $z$ direction in (b, c) and the $x$ direction in (d).
Thus the ground state is represented in the eigenbases of the $\sigma_i^z$ or $\sigma_i^x$ Pauli operators associated with each lattice site.
Training used contrastive divergence (CD) or persistent CD (PCD) with $\eta = 10^{-3}$, $B = 100$.
(a) Hamiltonian and sketch of the transverse-field Ising chain,
whose ground-state wave function $\psi(x)$ is the square root of the target distribution $p(x)$.
(b) Exact loss $\Delta_\theta$ (points)  and empirical loss $\tilde\Delta_\theta^{(S)}$ (solid lines) vs.\ autocorrelation time $\tau_\theta$ defined in~\eqref{eq:iact},
utilizing PCD (first column) or CD (last three columns), $\nCD = 10$ (second column) or $\nCD = 1$ (all other columns) and $\lvert S \rvert = 25\,000$ (first three columns) or $\lvert S \rvert = 500$ (fourth column) training samples,
measured in the $\sigma^z$ basis,
for several different values of the magnetic field $g$ (see left panel of each row).
Markers: $\Delta_\theta$ calculated from~\eqref{eq:LossDKL} with the filling color indicating the total correlation $\Ctot(\hat p_\theta)$ of the model distribution (see right colorbars),
and the border color and marker type indicating the number of hidden units $N$ (see second panel in first row).
Solid lines: $\tilde\Delta_\theta^{(S)}$ (see below Eq.~\eqref{eq:RBM:TrainingUpdate}),
partially masked under the $\Delta_\theta$ data and thus not visible.
Dashed lines: $\tau_\theta = 1$ (black), $\Delta_\theta = \Ctot(p)$ (red), $\Delta_\theta = c \, \tau_\theta^{-\alpha}$ (blue).
(c) $\Delta_\theta / \Ctot(p)$ vs.\ $\tau_\theta$ for various system sizes $M$ utilizing CD with $\nCD=1$, $N = 16$, $\lvert S \rvert = 25\,000$, $\sigma^z$ basis, and $g$ as indicated in each panel.
As a result of
rescaling
the loss $\Delta_\theta$ with the total correlation $\Ctot(p)$ of the target distribution, the learning curves collapse in the independent- and correlation-learning regimes.
Inset: Same data, but without the rescaling.
(d) $\Delta_\theta$ vs.\ $\tau_\theta$ for CD training in the $\sigma^x$ basis,
with $\lvert S \rvert = 25\,000$ samples
and
$\nCD = 1$.
Markers and dashed lines as in (b).
All curves correspond to averages over $5$ independent training runs.}
\label{fig:Tradeoff:TFIM}
\end{figure*}

\begin{figure*}
\centering
\includegraphics[scale=1]{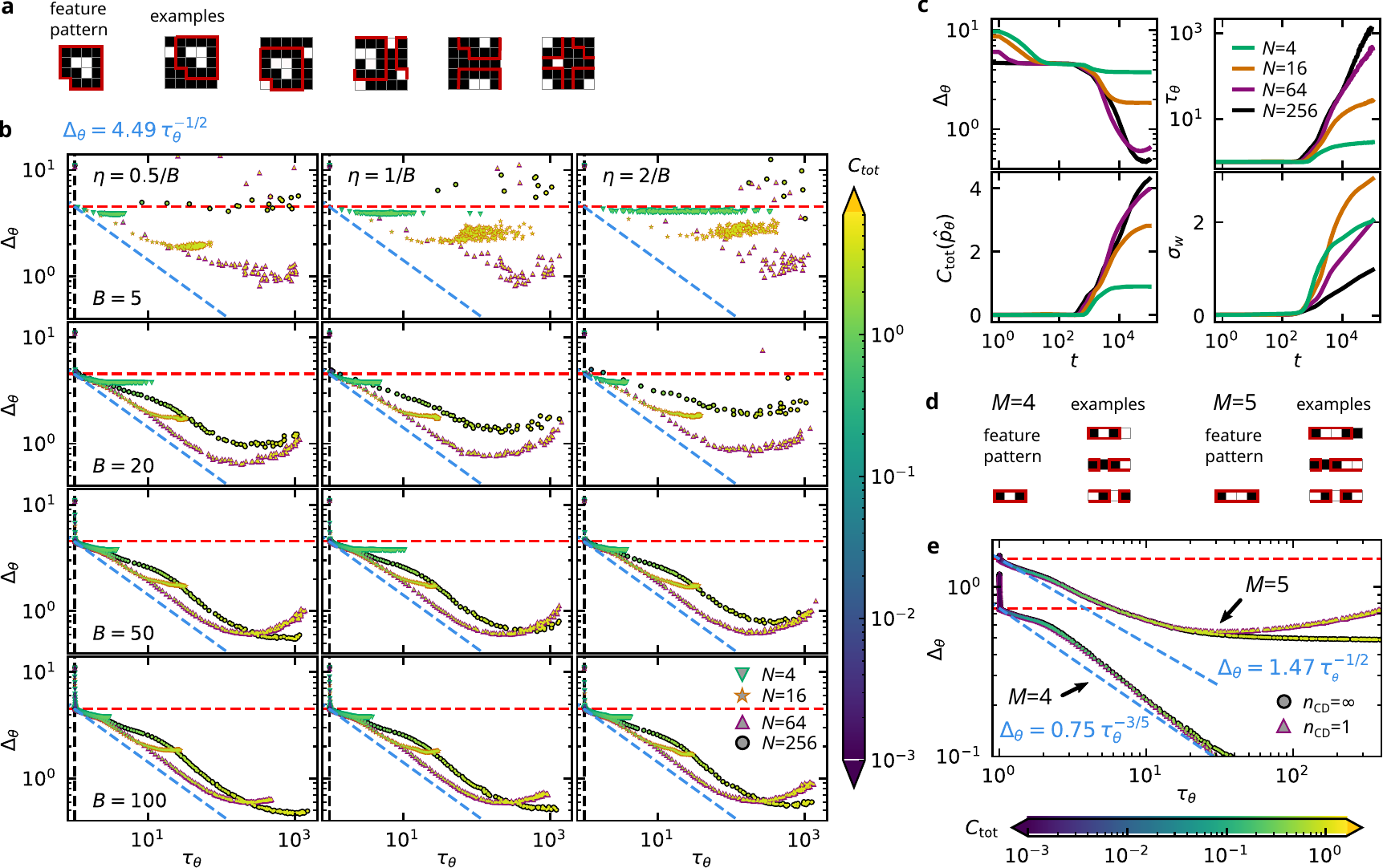}
\caption{RBM learning characteristics for a pattern recognition task.
(a) The target distribution consists of $\NV = 5 \times 5$ ``images'' subject to periodic boundary conditions and a fixed $15$-pixel ``hook'' pattern implanted at random locations,
where the remaining pixels are active (white) with probability $q = 0.1$.
(b) Exact loss~$\Delta_\theta$ vs.\ autocorrelation time~$\tau_\theta$ for RBMs with different numbers of hidden units $\NH$ (see legend),
trained on the distribution from (a)
using contrastive divergence of order $\nCD = 1$ with $\lvert S \rvert = 5000$ training samples and various values of the batch size $B$ (rows) and learning rate $\eta$ (columns).
Data points are averages over 5 independent runs.
(c) $\Delta_\theta$, $\tau_\theta$, total correlation $\Ctot(\hat p_\theta)$ of the model distribution, and the standard deviation of the weights $\sigma_w := (\frac{1}{\NV\NH - 1} \sum_{i,j} w_{ij}^{\,2})^{1/2}$
as a function of the training epoch $t$ for various $\NH$;
$\eta = 0.005$, $B = 100$ (cf.\ bottom left panel of (b)).
(d) Simplified model of $\NV = 1 \times 4$ or $\NV = 1 \times 5$ images with an implanted ``black-white(-white)-black'' pattern.
(e) $\Delta_\theta$ vs.\ $\tau_\theta$ for RBMs with $\NH = 2$ hidden units trained on the distributions from (d) using the full target distribution (i.e., $\lvert S \rvert = \infty$) and exact continuous-time gradient descent with either the full model distribution $\hat p_\theta(x, h)$ ($\nCD = \infty$) or contrastive divergence of order $\nCD = 1$.
Data points are averages over $100$ independent runs with different initial conditions.
In (b) and (e), fill colors indicate the total correlation $\Ctot(\hat p_\theta)$ of the model distribution (see colorbars),
border colors and marker types indicate the number of hidden units $N$ (see legends in bottom-right corners).}
\label{fig:Tradeoff:PH1}
\end{figure*}

\subsection*{Mechanism behind the learning stages}

\begin{figure*}
\centering
\includegraphics[scale=1]{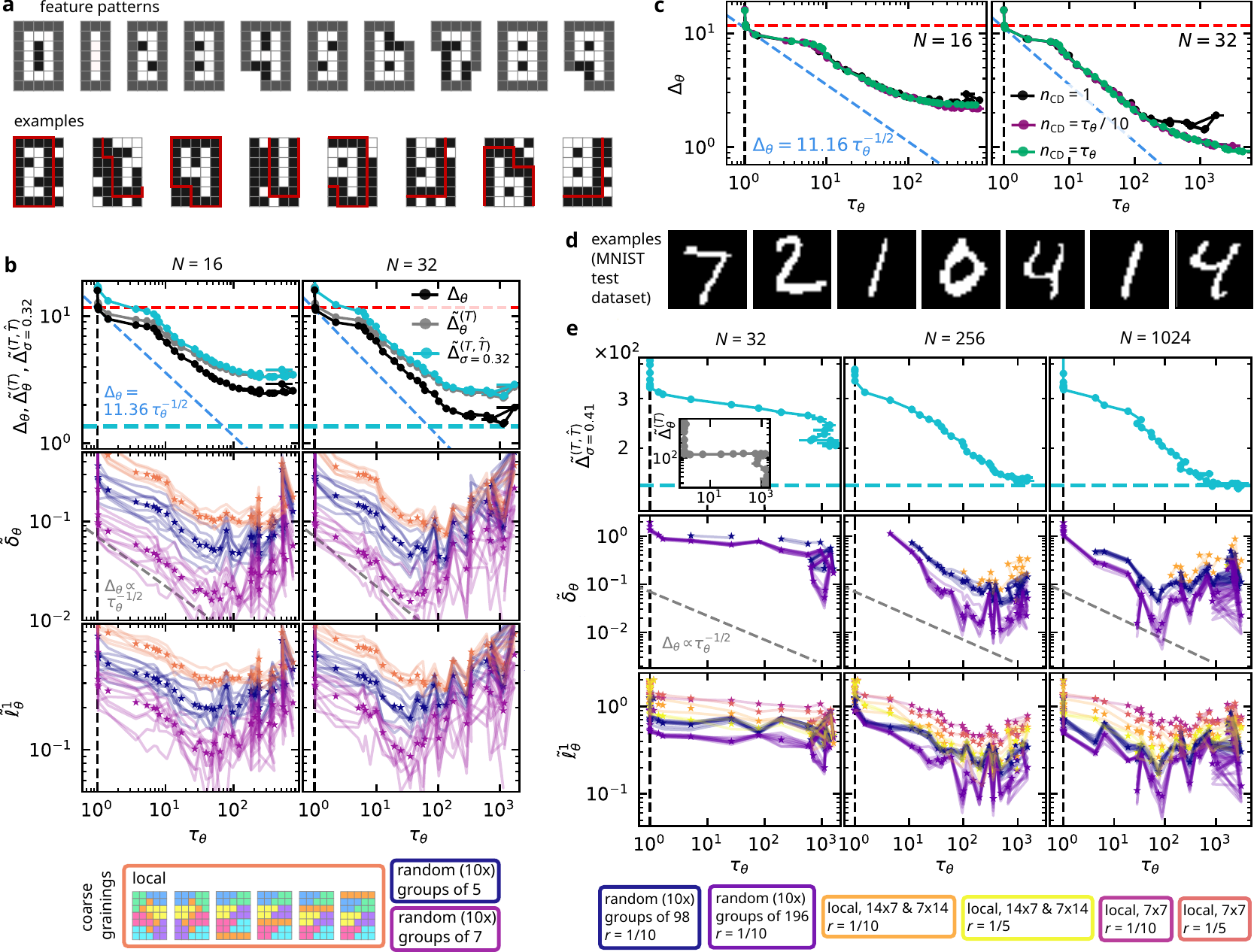}
\caption{Approximate RBM learning characteristics on digit images.
(a) Images of $\NV = 5\times 7$ pixels showing patterns of the digits 0 through 9 (selected uniformly) at a random location.
Gray pixels must either be
made black ($x_i = 0$) or be cut away by the image boundaries (see examples in the second row).
Pixels that are not part of the pattern are active (white) with probability $q = 0.1$.
The total number of such images is $40\,507\,353$.
(b) Various loss measures vs.\ autocorrelation time $\tau_\theta$ for $\NH = 16$ (left) and $\NH = 32$ (right) hidden units,
utilizing persistent contrastive divergence (PCD) with $\nCD = 1$, $\eta = 0.005$, $B = 100$ on $\lvert S \rvert = 50\,000$ training images.
Top: Exact loss $\Delta_\theta$ (black),
exact test error $\tilde\Delta_\theta^{(T)}$ (empirical loss for a test dataset $T$ of $\lvert T \rvert = 10\,000$ images, gray),
and Gaussian-smoothened empirical loss estimate $\tilde\Delta^{(T, \hat T)}_{\sigma}$ ($\lvert \hat T \rvert = 10^6$, $\sigma = 0.32$, cyan).
The cyan dashed line marks $\tilde\Delta^{(T, S)}_{\sigma=0.32} = 1.354$, the minimal Gaussian-smoothened loss estimate between the test and training datasets.
Middle: Empirical error $\tilde\delta_\theta$ using majority-rule ($r=1$) coarse-grainings of samples from the target and model distributions, partitioning pixels into local or random groups
(see main text for details).
Solid lines: results for individual partitions;
star markers: average of the solid lines of the same partitioning type (color, see legend).
Bottom: empirical error $\tilde\ell^1_\theta$ for the same coarse-grainings.
(c) $\Delta_\theta$ vs.\ $\tau_\theta$ for $\NH = 16$ (left) and $\NH = 32$ (right) hidden units,
utilizing persistent contrastive divergence (PCD) with fixed ($\nCD = 1$) or adaptive ($\nCD \propto \tau_\theta$) approximation order;
other hyperparameters as in (b).
(d) Examples from the MNIST dataset,
which comprises images of $\NV = 28\times 28$ pixels showing handwritten digits.
(e) Similar to (b), but for the MNIST dataset and PCD training with $\nCD = 1$, $\eta = 10^{-4}$, $B = 100$, $\lvert S \rvert = 60\,000$, $\lvert T \rvert = 10\,000$, $\lvert \hat T \rvert = 10^6$, $\sigma = 0.41$, and $\tilde\Delta^{(T, S)}_{\sigma=0.41} = 147.4$.
Missing data points correspond to $\tilde\delta_\theta =  \infty$ and/or unrealiable $\tau_\theta$ estimates.}
\label{fig:Tradeoff:Applications}
\end{figure*}

With no
specific knowledge about the target distribution,
it is natural to initialize the RBM parameters $\theta = (\theta_k) = (w_{ij}, a_i, b_j)$ randomly.
Moreover, the initial values should be sufficiently small
so that any spurious correlations arising from the initialization are much smaller than the actual correlations in the target distribution and can be overcome within a few training steps.
In the examples from Figs.~\ref{fig:Tradeoff:TFIM}--\ref{fig:Tradeoff:Applications},
we draw the initial $\theta_k$ independently from a normal distribution $\mc N(\mu, \sigma)$ of mean $\mu$ and standard deviation $\sigma$,
namely
$w_{ij} \sim \mc N(0, 10^{-2})$ and
$a_i, b_j \sim \mc N(0, 10^{-1})$
unless stated otherwise.
A brief exploration of other initialization schemes,
including Hinton's proposal \cite{Hinton:2012pgt} and examples with significant (spurious) correlations,
can be found in Supplementary Note~3.
In Figs.~\ref{fig:Tradeoff:TFIM} and~\ref{fig:Tradeoff:PH1},
the experiments were repeated for $5$ independent runs for each hyperparameter configuration,
and the displayed data are averages over those runs at fixed training epoch $t$.
No error bars are shown in these figures for clarity,
but the spread of the point clouds typically serves as a decent visualization of the uncertainty.
We also highlight that important information for the ensuing discussion is encoded in the coloring of the data points.
Particularly,
both the filling color and the border color convey correlation characteristics and hyperparameter dependencies as indicated in the legends and figure captions.

We now sketch
how the three learning regimes and the tradeoff relation arise.
Intuitively, the
origin of the accuracy--efficiency tradeoff
can be understood as follows:
To improve the model representation $\hat p_\theta(x)$ of the target distribution $p(x)$,
correlations of $p(x)$ between the different visible units $x_i$ have to be incorporated into $\hat p_\theta(x)$.
Since correlations between visible units are mediated by the hidden units in the RBM model~\eqref{eq:RBM:pHX},
this inevitably increases the correlation between subsequent samples in the Markov chain~\eqref{eq:MarkovChainXH}
and thus leads to larger autocorrelation times $\tau_\theta$ in~\eqref{eq:iact}.
Nevertheless, the detailed relationship between $\Delta_\theta$ and $\tau_\theta$ and its remarkable structural universality turn out to be more subtle as discussed in the following.

In the \emph{independent-learning regime},
which constitutes
the first stage of the natural learning dynamics,
the loss $\Delta_\theta$ is actually reduced without any significant increase of the autocorrelation time $\tau_\theta$.
Hence the RBM picks up aspects of the target distribution whilst preserving independence of its visible units.
The minimal loss $\Delta_\theta$ that can be achieved with a product distribution of independent units $x_i$ is given by the \emph{total correlation} \cite{Watanabe:1960ita}
\begin{equation}
\label{eq:Ctot}
	\Ctot(p) := \sum_x p(x) \log \frac{p(x)}{p_1(x_1) \cdots p_{\NV}(x_{\NV})}
\end{equation}
of the target distribution.
This quantity is thus
the
KL
divergence (cf.\ Eq.~\eqref{eq:LossDKL}) from the product of marginal distributions $p_i(x_i)$ to the joint distribution $p(x) = p(x_1, \ldots, x_{\NV})$.
It can be understood as a multivariate analog of mutual information.
For an arbitrary product distribution $\hat p(x) := \prod_i \hat p_i(x_i)$,
we have
$\DKL(p || \hat p) = \Ctot(p) + \sum_i \DKL(p_i || \hat p_i) \geq \Ctot(p)$ (see Supplementary Note~5).
Hence $\Ctot(p)$ indeed lower-bounds the loss $\Delta_\theta$ for independent units.

The value of $\Ctot(p)$ is indicated by the red dashed lines in Figs.~\ref{fig:Sketch3Regimes}--\ref{fig:Tradeoff:Applications},
and indeed marks the end of the independent-learning regime
as defined by $\tau_\theta \simeq 1$ in Figs.~\ref{fig:Tradeoff:TFIM}--\ref{fig:Tradeoff:Applications}.
As a consequence, we can identify the constant $c$ from the tradeoff relation~\eqref{eq:Tradeoff},
which bounds $\Delta_\theta$ from below at $\tau_\theta = 1$
(see also Methods),
with the total correlation $\Ctot(p)$ of the target distribution, $c \simeq \Ctot(p)$,
as illustrated by the intersection of the red ($\Delta_\theta = \Ctot(p)$), blue ($\Delta_\theta = c \, \tau_\theta^{\,-\alpha})$, and black ($\tau_\theta = 1$) dashed lines in Figs.~\ref{fig:Sketch3Regimes}--\ref{fig:Tradeoff:Applications}.

A closer inspection of the total correlation $\Ctot(\hat p_\theta)$ of the \emph{model} distribution,
encoded by the color gradients in Figs.~\ref{fig:Tradeoff:TFIM}--\ref{fig:Tradeoff:PH1},
confirms that no significant correlations between the RBM's visible units build up as long as $\tau_\theta \simeq 1$,
providing further justification for labeling this stage as the ``independent-learning'' regime.
The time spent in this regime can be reduced by adjusting the biases $a_i$ to the activation frequencies of the visible units in the training data as suggested by Hinton \cite{Hinton:2012pgt} (see also Supplementary Note~3).

The
independent-learning regime is thus characterized by $\tau_\theta \simeq 1$ and $\Delta_\theta \gtrsim \Ctot(p)$.
As soon as $\Delta_\theta$ falls below $\Ctot(p)$,
the RBM enters the \emph{correlation-learning regime}
and starts to exhibit noticeable dependencies between its visible units,
accompanied by an increase of $\tau_\theta$.
This regime is characterized by $\Delta_\theta \lesssim \Ctot(p)$ and $\frac{\partial \tau_\theta}{\partial \Delta_\theta} < 0$,
meaning that $\tau_\theta$ grows as $\Delta_\theta$ decreases.
Quantitatively (cf.\ Figs.~\ref{fig:Tradeoff:TFIM}b--d, \ref{fig:Tradeoff:PH1}b,e, \ref{fig:Tradeoff:Applications}b,c),
we find that the functional dependence between $\Delta_\theta$ and $\tau_\theta$ is
(piecewise) power-law-like and often closely follows the lower bound provided by the
tradeoff relation~\eqref{eq:Tradeoff}.

In most of our examples,
the exponent $\alpha$
turns out
to be well approximated by $\alpha \simeq \frac{1}{2}$.
The notable exception is the example
in Fig.~\ref{fig:Tradeoff:TFIM}d of TFIC ground-state tomography in the $\sigma^x$ basis
(but not the $\sigma^z$ basis; see figure caption for details),
where a value of $\alpha \approx 6 \ldots 8$ seems more appropriate.
Roughly speaking,
$\alpha$ quantifies how efficiently the prevailing correlations in the target distribution $p(x)$ can be encoded in the RBM model $\hat p_\theta(x)$.
A larger value of $\alpha$ implies that the tradeoff~\eqref{eq:Tradeoff} is less severe,
indicating a closer structural similarity of $p(x)$ to the model family $\hat p_\theta(x)$.

The relationship between accuracy and efficiency in the correlation-learning regime turns out to be remarkably stable against variations of the architecture or the training details,
suggesting that it is indeed an intrinsic limitation of the RBM model
whose qualitative details are essentially determined by the target distribution.
First,
as long as training is stable,
the $\Delta_\theta$--$\tau_\theta$ learning trajectories
are almost
independent of further hyperparameters such as the number of training samples $\lvert S \rvert$, the minibatch size $B$, or the learning rate $\eta$.
This is illustrated in Fig.~\ref{fig:Tradeoff:PH1}b
(see Supplementary Note~5 for further examples),
which also visualizes how training becomes unstable if $\eta$ or $B$ become too small,
leading to underperforming machines with $(\tau_\theta, \Delta_\theta)$
further away from the global bound~\eqref{eq:Tradeoff}.
Second,
changing the approximation of the model averages in~\eqref{eq:RBM:TrainingUpdate} does not affect the relation between $\Delta_\theta$ and $\tau_\theta$.
In fact, approximation schemes
which achieve a smaller loss $\Delta_\theta$
increase the autocorrelation time $\tau_\theta$ 
in accordance with the tradeoff~\eqref{eq:Tradeoff}.
This is exemplified by variations in the order ($\nCD$) and the initialization (CD vs.\ PCD) of the training chains~\eqref{eq:MarkovChainXH} in Fig.~\ref{fig:Tradeoff:TFIM}b.
Third,
as long as the loss is sufficiently
above
the expressivity threshold (see below),
the relationship between $\Delta_\theta$ and $\tau_\theta$ is largely insensitive to the number of hidden units $N$ (see Figs.~\ref{fig:Tradeoff:TFIM}b,d, \ref{fig:Tradeoff:PH1}b, \ref{fig:Tradeoff:Applications}b,c).
Fourth,
the learning characteristics appear to be intrinsic to the problem \emph{type},
but not its \emph{size} if a natural scaling for the number of visible units $\NV$ exists.
To this end, we consider the TFIC example and vary the number of lattice sites $\NV$ in Fig.~\ref{fig:Tradeoff:TFIM}c.
While this changes the total correlation $\Ctot(p)$ of the target distribution,
the rescaled curves of $\Delta_\theta / \Ctot(p)$ vs.\ $\tau_\theta$ collapse almost perfectly onto
a single universal curve in the independent- and correlation-learning regimes.

The end of the correlation-learning regime and the crossover into the \emph{degradation regime} is influenced by various (hyper)parameters.
An absolute limit for the minimal value of $\Delta_\theta$
results from the class of distributions that can be represented by the RBM.
This ``expressivity''
is controlled by the number of hidden units $\NH$.
For sufficiently large $\NH$, the RBM model can approximate any target distribution with arbitrary accuracy \cite{Younes:1996sbm, LeRoux:2008rpr, Montufar:2017hma, Montufar:2018rbm};
hence there is no absolute
minimum for $\Delta_\theta$ in principle.
In practice, however, the number of hidden units is limited by the available computational resources.
Note that the scaling of this expressivity threshold is analyzed in some detail in Ref.~\cite{Sehayek:2019lsq} for the TFIC example (cf.\ Fig.~\ref{fig:Tradeoff:TFIM}).

Ceasing
accuracy improvement due to limited expressivity is
exemplified
by Fig.~\ref{fig:Tradeoff:PH1}b 
in the stable regime ($B \gtrsim 50$),
where we note that the achievable minimal loss decreases significantly from $\NH = 4$ to $16$ to $64$
(the same behavior can also be observed in Fig.~\ref{fig:Tradeoff:Applications}b,c).
Employing even more hidden units, however, does not facilitate any significant gain in accuracy,
and the learning characteristics for $\NH = 256$ in Fig.~\ref{fig:Tradeoff:PH1}b actually signal slightly worse performance in terms of the accuracy--efficiency tradeoff,
i.e., a larger offset from the global lower bound (blue dashed line).

If $N$ is sufficiently large,
the approximations leading to a bias of the (exact) update step~\eqref{eq:RBM:TrainingUpdate} will usually take over eventually 
and lead into the degradation regime even if the expressivity threshold has not yet been reached.

The first of those approximations is the use of the empirical distribution $\tilde p(x; S)$
in lieu of the unknown true target distribution $p(x)$.
This may result in overfitting,
a phenomenon common to many machine-learning approaches:
The RBM may pick up finite-size artifacts of $\tilde p(x; S)$,
particularly when the
resolution of genuine features in the model distribution approaches the resolution of those features in the empirical distribution.
Overfitting is the primary reason for degradation in the fourth column of Fig.~\ref{fig:Tradeoff:TFIM}b,
where the size of the training dataset $\lvert S \rvert$ is rather small.
Comparing the training error $\tilde\Delta_\theta^{(S)}$ (solid lines, see below~\eqref{eq:RBM:TrainingUpdate})
with the test error $\Delta_\theta$ (data points, see Eq.~\eqref{eq:LossDKL}),
we observe that the former continues to decrease even though the latter actually increases.

In the first three columns of Fig.~\ref{fig:Tradeoff:TFIM}b,
by contrast,
$\tilde\Delta_\theta^{(S)}$ usually follows $\Delta_\theta$ closely (thus the solid lines are often hidden behind the data points).
Here, degradation is due to the second limiting approximation of the update step~\eqref{eq:RBM:TrainingUpdate},
namely the replacement of averages over the model distribution $\hat p_\theta(x, h)$ by Markov-chain samples~\eqref{eq:MarkovChainXH}.
In fact, this is directly related to the definition of $\tau_\theta$
because larger values imply that the chain~\eqref{eq:MarkovChainXH} needs to be run for a longer time in order to obtain an effectively independent sample (see below Eq.~\eqref{eq:iact}).
Indeed,
smaller losses can be achieved for larger $\nCD$
(second vs.\ third column).
Similarly, at fixed $\nCD$,
PCD can reach higher accuracies than CD (first vs.\ third column; see also Supplementary Note~5).

Finally, we turn to the smallest example from Fig.~\ref{fig:Tradeoff:PH1}d--e.
In this case, we can directly integrate the continuous-time ($\eta = 0$) update equations~\eqref{eq:RBM:TrainingUpdate} with the full target distribution $p(x)$ (i.e., $\lvert S \rvert = \infty$)
and the exact model distribution $\hat p_\theta(x, h)$ (i.e., $\nCD = \infty$) for RBMs with $\NH = 2$ hidden units (see also Supplementary Note~1).
We again observe a power-law tradeoff between $\Delta_\theta$ and $\tau_\theta$ with $\alpha \simeq \frac{1}{3} \ldots \frac{3}{5}$,
limited by the machine's expressivity in the $M = 5$, but not in the $M = 4$ case.
Moreover,
by averaging over $\hat p_\theta^{(1)}(x, h) := \hat p_\theta(h \given x) \sum_{x',h'} \hat p_\theta(x \given h') \hat p_\theta(h' \given x') p(x')$ instead of $\hat p_\theta(x, h)$ in~\eqref{eq:RBM:TrainingUpdate},
we can adopt the exact CD update of order $\nCD = 1$.
This reintroduces the correlation bias into the updates and indeed leads to stronger deviations from the power-law behavior for $M = 5$, with increasing $\Delta_\theta$ in the degradation regime.

\subsection*{Towards applications}

All examples discussed so far (Figs.~\ref{fig:Tradeoff:TFIM}, \ref{fig:Tradeoff:PH1} and~\ref{fig:Tradeoff:Applications}a--c) involved only a small number of visible units $\NV$
so that the accuracy measure $\Delta_\theta$ could be evaluated numerically exactly.
In practice, this is impossible because neither the target distribution $p(x)$ nor the model distribution $\hat p_\theta(x)$ are directly accessible.
In the following, we will sketch how learning characteristics and the accuracy--efficiency tradeoff can be analyzed approximately in applications
and apply the ideas, in particular, to the MNIST dataset \cite{LeCun:YYYYmdh}
as a standard machine-learning benchmark of larger problem size (see Fig.~\ref{fig:Tradeoff:Applications}d,e).

To approximate the accuracy measure $\Delta_\theta$,
the target distribution $p(x)$ is usually replaced by the empirical distribution $\pEmp(x; T)$ for a (multi)set of test samples $T$ (independent of the training samples $S$).
If both $\NV$ and $\NH$ become large,
$\hat p_\theta(x)$ must be approximated by an empirical counterpart as well.
To this end,
a collection of independent samples from $\hat p_\theta(x)$ is needed.
Typically, it will be generated approximately by Markov chains~\eqref{eq:MarkovChainXH},
which directly leads back to the autocorrelation time $\tau_\theta$ from~\eqref{eq:iact} as a measure for the number of steps required in~\eqref{eq:MarkovChainXH} to obtain an effectively independent sample.

Estimating $\tau_\theta$,
in turn,
should remain feasible along the lines outlined in Methods even if $\NV$ and $\NH$ are large.
To be precise,
if it turns out to be impossible in practice to reliably estimate $\tau_\theta$,
then any conclusions about the model distribution $\hat p_\theta(x)$ drawn from Markov chains like~\eqref{eq:MarkovChainXH} are equally unreliable.
In other words,
if $\tau_\theta$ (or, more generally, the integrated autocorrelation time of the observable of interest)
cannot be computed,
the trained model itself becomes useless as a statistical model of the target distribution.
A particular challenge are metastabilities where the chains spend large amounts of time in a local regime of the configuration space and only rarely transition between those regimes.
These can be caused, for instance, by a multimodal structure of the target distribution.
If undetected, those metastabilities can lead to vastly underestimated autocorrelation times.

Once a set of (approximately) independent samples $\hat T$ from $\hat p_\theta(x)$ is available,
the KL divergence $\DKL(\pEmp(\,\cdot\,; T) || \pEmp(\,\cdot\,; \hat T))$
can serve as a proxy for $\Delta_\theta$ in principle.
In practice, however, this approach will not be viable because this proxy diverges whenever there is a sample $\tilde x$ in $T$ which is not found in $\hat T$,
meaning that the sample size required for $\hat T$ will often be out of reach.

We suggest two alternative approaches to mitigate this problem.
First, we consider smoothening the empirical model distribution $\pEmp(x; \hat T)$ by convolving it with a Gaussian kernel $k(x; \mu, \sigma) := N_\sigma^{-1} \e^{-(x-\mu)^2 / 2 \sigma^2}$,
where $N_\sigma := \sum_{d=0}^{\NV} \binom{\NV}{d} \e^{-d / 2 \sigma^2}$,
leading to $\pEmp_\sigma(x, \hat T) := \frac{1}{\lvert \hat T \rvert} \sum_{\hat x \in \hat T} k(x; \hat x, \sigma)$.
The KL divergence $\tilde\Delta^{(T, \hat T)}_{\sigma} := \DKL(\pEmp(\,\cdot\,; T) || \pEmp_\sigma(\,\cdot\,; \hat T))$ then approximates $\Delta_\theta$, where $\sigma$ is chosen so as to make $\tilde\Delta^{(T, S)}_\sigma$ minimal, i.e., when using the training data $S$ as the empirical model distribution \cite{Desjardins:2010tmc} (see also Supplementary Note~2).
As shown in the first row of Fig.~\ref{fig:Tradeoff:Applications}b,
$\Delta^{(T, \hat T)}_\sigma$ reproduces essentially the same behavior as $\Delta_\theta$ and $\tilde\Delta^{(T)}_\theta$.

Second, we propose coarse-graining the samples in $T$ and $\hat T$,
such that every $\tilde x = (\tilde x_1, \ldots, \tilde x_{\NV}) \in T, \hat T$ is mapped to a new configuration $\tilde y = (\tilde y_1, \ldots, \tilde y_L)$ with $\tilde y_l \in \{ 0, 1 \}$ and $L < \NV$.
Denoting the resulting multisets of reduced configurations by $T'$ and $\hat T'$,
we then consider the KL divergence $\tilde\delta_\theta := \DKL(\pEmp(\,\cdot\,; T')||\pEmp(\,\cdot\,; \hat T'))$ of the associated empirical distributions as a qualitative approximation of $\Delta_\theta$.
To be specific, in Fig.~\ref{fig:Tradeoff:Applications},
we employ a weighted majority rule for coarse graining using random or local partitions of the visible units into $L$ subsets,
such that $\tilde y_l = 1$ if a fraction of $r$ or more units in the $l$th subset is active (see Methods for details).

While some of the quantitative details are inevitably lost as a result of the coarse graining,
the results in Fig.~\ref{fig:Tradeoff:Applications}b show that the accuracy measure $\tilde\delta_\theta$ still conveys similar learning characteristics as the exact loss $\Delta_\theta$.
Remarkably, even the same exponent $\alpha \simeq \frac{1}{2}$ is found to describe the tradeoff between $\tilde\delta_\theta$ and $\tau_\theta$ in the correlation-learning regime.
On the other hand, the coarse-grained loss $\tilde\delta_\theta$ appears to deteriorate somewhat prematurely,
especially for the random coarse grainings,
indicating that late improvements of $\Delta_\theta$ involve finer, presumably local correlations
that cannot be captured by $\tilde\delta_\theta$ in these cases.

Furthermore, we also consider the $L^1$ distance $\tilde\ell^1_\theta := \sum_x \lvert \pEmp(x; T') - \pEmp(x; \hat T') \rvert$ between the reduced empirical distributions as an accuracy measure.
Its advantage is that---unlike $\tilde\delta_\theta$---it does not suffer from divergences when $T' \nsubseteq \hat T'$ (cf.\ Fig.~\ref{fig:Tradeoff:Applications}e in particular).
As shown in Fig.~\ref{fig:Tradeoff:Applications}b and~e,
the $\tilde\ell^1_\theta$--$\tau_\theta$ curves qualitatively agree with their $\tilde\delta_\theta$--$\tau_\theta$ counterparts and can thus serve as a more stable way to monitor the tradeoff in case of smaller sample sizes.

Inspecting the learning characteristics in the MNIST example from Fig.~\ref{fig:Tradeoff:Applications}e,
we observe that the relationship between the accuracy measures $\tilde\Delta^{(T, \hat T)}_\sigma$, $\tilde\delta_\theta$, $\tilde\ell^1_\theta$ and the efficiency measure $\tau_\theta$ are qualitatively similar as in the simpler example in Fig.~\ref{fig:Tradeoff:Applications}b,
especially for the more expressive RBMs with $\NH \geq 256$.
Notably, we find an initial regime with decreasing $\tilde\Delta^{(T, \hat T)}_\sigma$ and $\tilde\ell^1_\theta$ at $\tau_\theta = 1$ ($\tilde\delta_\theta = \infty$ here due to the aforementioned undersampling problem),
followed by an approximately power-law-like tradeoff between accuracy and efficiency,
and finally ceasing improvement ($\tilde\Delta^{(T, \hat T)}_\sigma$) or deterioration ($\tilde\delta_\theta$, $\tilde\ell^1_\theta$) at increasing $\tau_\theta$.
For $\NH = 32$, by contrast, the RBM accuracy does not improve much beyond the independent-learning threshold,
except for somewhat unstable fluctuations at very late training stages.
Hence we expect that the same tradeoff mechanism identified in the small-scale examples from Figs.~\ref{fig:Tradeoff:TFIM} through~\ref{fig:Tradeoff:Applications}a--c
also governs the behavior of more realistic, large-scale learning problems.

Altogether,
our present results suggest a couple of approaches to monitor the accuracy and efficiency in applications with large input dimension $M$.
First, we propose estimating the autocorrelation time $\tau_\theta$
at selected epochs during training
and stop when it exceeds the threshold set by the available evaluation resources in the intended
use case.
Second,
it may be helpful to train RBMs with smaller numbers of hidden units $\NH$
so that the test error $\tilde\Delta_\theta^{(T)}$ can be evaluated exactly (see also Methods),
even though those small-$\NH$ machines will typically not reach the desired accuracies.
Since the onset of the correlation-learning regime and the subsequent initial progression are essentially independent of $\NH$,
the relationship between $\tilde\Delta_\theta^{(T)}$ and $\tau_\theta$ for small $\NH$ can provide an intuition and perhaps even a cautious extrapolation of the behavior for larger $\NH$.
Third,
empirical accuracy measures such as $\tilde\Delta^{(T, \hat T)}$ , $\tilde\delta_\theta$ and $\tilde\ell^1_\theta$
can assure that the machine is still learning
and possibly even map out the beginning of the degradation regime.
Fourth,
estimates of $\tau_\theta$ can be naturally obtained \emph{en passant} when using the PCD algorithm.
These estimates can then be employed to adapt the length $\nCD$ of the Markov chains~\eqref{eq:MarkovChainXH} to the current level of correlations when approximating the model averages in~\eqref{eq:RBM:TrainingUpdate}.
While we leave a detailed analysis of the resulting ``adaptive PCD'' algorithm for future work,
preliminary results (see Fig.~\ref{fig:Tradeoff:Applications}c) suggest that one can indeed reach better accuracies this way,
while the tradeoff~\eqref{eq:Tradeoff} remains valid.

\section*{Discussion}

In summary,
the accuracy--efficiency tradeoff is an inherent limitation of the RBM architecture and its reliance on Gibbs sampling~\eqref{eq:MarkovChainXH} to assess the model distribution $\hat p_\theta(x)$.
Depending on the eventual application of the trained model,
this limitation should already be taken into account when planning and performing training:
Aiming at higher accuracy implies that more resources will be required also in the production stage to evaluate and employ the trained model in an unbiased fashion.

Not least, the tradeoff directly affects the training process itself.
It is well known that common training algorithms like contrastive divergence and its variants are biased \cite{Hinton:2002tpe, CarreiraPerpinan:2005cdl}
and that the bias increases with the magnitude of the weights \cite{Bengio:2009jgc, Fischer:2010ead}.
Hence there exists an optimal stopping time for training
at which the accuracy becomes maximal,
but unfortunately, no simple criterion in terms of accessible quantities is known to determine
this stopping time \cite{Fischer:2010ead, Schulz:2010icr}.
Approximate test errors like $\tilde\Delta^{(T, \hat T)}_\sigma$, $\tilde\delta_\theta$ or $\tilde\ell^1_\theta$ can provide a rough estimate for when deterioration sets in,
but are insensitive to finer details by construction.
By contrast,
taking the reconstruction error as a measure for the model accuracy,
which is still not uncommon since it is easily accessible,
is downright detrimental from a sampling-efficiency point of view
because it decreases with increasing correlations between samples.
Since it is not correlated with the actual loss either \cite{Fischer:2010ead},
the reconstruction error should rather be regarded as an efficiency measure
(with larger ``error'' indicating higher efficiency).

The aforementioned fact that the magnitude of the weights is closely related to the autocorrelation time $\tau_\theta$
(see also Supplementary Note~5) provides a dynamical understanding of the bias
in the sense that larger $\tau_\theta$ calls for more steps in the Markov chain~\eqref{eq:MarkovChainXH} to obtain an effectively independent sample.
Similar conclusions have been drawn from studies of the mixing time of RBM Gibbs samplers \cite{Bengio:2009jgc, Fischer:2015bcr, Tosh:2016mra, Decelle:2021ene}.
The mixing time quantifies how many steps in~\eqref{eq:MarkovChainXH} are necessary to reach the stationary distribution $\hat p_\theta(x)$ from an arbitrary initial distribution for $x^{(0)}$.
In CD training, where $x^{(0)}$ is taken from the training data (meaning that it is a sample drawn from $p(x)$ by assumption),
it is
particularly relevant for the early training stages when $\hat p_\theta(x)$ is possibly far away from the target.
For analyzing a trained model,
by contrast,
the mixing time is less important because it only provides a constant offset to the sampling efficiency by quantifying the burn-in steps in~\eqref{eq:MarkovChainXH},
i.e., the number of samples to discard until the stationary regime is reached,
whereafter one will start recording samples to actually assess $\hat p_\theta(x)$.
Similarly, correlations in the PCD update steps
are better described by autocorrelation times like $\tau_\theta$,
at least if the learning rate is sufficiently small so that the Markov chains can be considered to operate in the stationary regime throughout training,
and the same applies to ordinary CD updates at later training stages.

There are a variety of proposals to modify the sampling process so that correlations between subsequent samples in
an appropriate analog of~\eqref{eq:MarkovChainXH} are reduced,
including
the above-sketched PCD extension with $\tau_\theta$-adaptive order of the Markov-chain sampling (see also Fig.~\ref{fig:Tradeoff:Applications}c),
parallel tempering \cite{Salakhutdinov:2009lmr, Desjardins:2010tmc},
mode-assisted training \cite{Manukian:2020mau},
or occasional Metropolis-Hastings updates \cite{Bruegge:2013fst, Roussel:2021bdp}.
However, these adaptations come with their own caveats
and the extent to which correlations are reduced may depend strongly on the setting \cite{Fischer:2015bcr, Roussel:2021bdp}.
Moreover, the computational complexity of these methods is usually higher
because additional substeps are necessary to produce a new Markov-chain sample.
While a detailed quantitative analysis is missing,
the overall evaluation efficiency (e.g., the required computational resources) will presumably not be improved in general \cite{Long:2010rbm},
and probably the only remedy to circumvent the sampling problem could be novel computing hardware such as neuromorphic chips \cite{Petrovici:2016sis, Kungl:2019ape, Czischek:2019ssn, Czischek:2022snc, Klassert:2021vlq}, ``memcomputing machines'' \cite{Manukian:2019adl}, or quantum annealers \cite{Adachi:2015aqa, Benedetti:2016eet}.

For a more comprehensive understanding of the tradeoff mechanism,
it would be desirable to elucidate the role of the exponent $\alpha$ in~\eqref{eq:Tradeoff} and how it relates to properties of the target distribution $p(x)$.
As discussed above, $\alpha$ roughly quantifies how apt the RBM architecture is to represent $p(x)$,
with larger values of $\alpha$ indicating better suitability.
A related question is what distributions can be represented efficiently by RBMs in terms of the required number of hidden units \cite{Younes:1996sbm, Martens:2013rer, Montufar:2017hma}.
Besides the number of ``active'' states,
symmetries that make it possible to represent the correlations between various visible units with fewer hidden units could play an important role in affecting $\alpha$ (see also Supplementary Note~5).
Furthermore,
observing the marked transition from independent to correlation learning,
one may naturally wonder whether there exists a hierarchy of how and when correlations are adopted during the correlation-learning regime \cite{Amari:2001igh, LeRoux:2011lgma, Lin:2017wdd, Montufar:2017hma, Saxe:2019mts},
particularly when $\alpha$ is ambiguous (e.g., in Fig.~\ref{fig:Tradeoff:TFIM}d; see also Supplementary Note~5).
In any case, it is
remarkable
that in most of the examples we explored,
$\alpha$
turns out
to be approximately $\frac{1}{2}$,
particularly
at the initial stage
of the correlation-learning regime.
Whether this is a coincidence or a hint at some deeper universality principle is an intriguing open question.

\section*{Methods}

\subsection*{Conditional RBM distributions}

The approach of using alternating Gibbs sampling of visible and hidden units via Markov chains of the form~\eqref{eq:MarkovChainXH}
is viable in practice only due to the bipartite structure of the RBM with direct coupling exclusively between one visible and one hidden unit.
Consequently, the visible units are conditionally independent given the hidden ones and vice versa,
e.g.,
$\hat p_\theta(h \given x) = \prod_j \hat p_\theta(h_j \given x)$ with
\begin{equation}
\label{eq:RBM:pXGivenH}
	\hat p_\theta(h_j \given x) = \frac{ \e^{ (\sum_i w_{ij} x_i + b_j) h_j } }{ 1 + \e^{ \sum_i w_{ij} x_i + b_j } } \,,
\end{equation}
and similarly $\hat p_\theta(x \given h)$
can be obtained by replacing $x_i \leftrightarrow h_j$ and $a_i \leftrightarrow b_j$
and by summing over $i$ in the exponents and taking the product over $j$.
Sampling from $\hat p_\theta(h \given x)$ and $\hat p_\theta(x \given h)$ is thus of polynomial complexity in the number of units and can be carried out efficiently.
Likewise, this explains why the first average on the right-hand side of~\eqref{eq:RBM:TrainingUpdate} with $\tilde p(x; S)$ in lieu of $p(x)$ (sometimes called the ``data average;'' see also below Eq.~\eqref{eq:RBM:TrainingUpdate}) can be readily evaluated.
For $\theta_k = w_{ij}$, for example, one finds
\begin{equation}
\label{eq:RBM:DatAvgW}
	\left\< \tfrac{\partial E_\theta(x, h)}{\partial w_{ij}} \right\>_{\! \hat p_\theta(h \given x) \tilde p(x; S)}
		= -\frac{1}{\lvert S \rvert} \sum_{x \in S} x_i \, \hat p_\theta(h_j = 1 \given x) \,,
\end{equation}
and similarly for $a_i$ and $b_j$.

The variability of samples obtained from those conditional distributions can be assessed in terms of their Shannon entropy,
defined for an arbitrary probability distribution $p(x)$ as $S(p) := -\sum_x p(x) \, \log p(x)$.
Specifically,
\begin{equation}
\label{eq:pHGivenXEntropy}
\begin{aligned}
	& S(\hat p_\theta(h \given x)) \\
		&\;= \sum_j \left[ \log\!\left( 1 + \e^{\sum_{i} w_{ij} x_i + b_j} \right) - \frac{ \sum_i w_{ij} x_i + b_j }{ 1 + \e^{-\sum_i w_{ij} x_i - b_j} } \right] ,
\end{aligned}
\end{equation}
and, again, similarly for $\hat p_\theta(x \given h)$.
The entropy is maximal for the uniform distribution with $\theta_k = 0$ for all parameters.
It remains large as long as the  $\theta_k$'s are small in magnitude
and tends to decrease
towards zero as $\lvert \theta_k \rvert$ increases
unless there is a special fine-tuning for specific configurations $h$ that leads to exact cancelations.
Over multiple steps of the Markov chain~\eqref{eq:MarkovChainXH},
the samples will thus generically show more variability for small weights,
whereas they develop stronger correlations as the weights grow \cite{Bengio:2009jgc, Fischer:2010ead} (see also Supplementary Note~5).

\subsection*{Details on $\Delta_\theta$, $\Ctot(\hat p_\theta)$ and related quantities}

The
measure of accuracy $\Delta_\theta$ (exact loss, ideal test error) is calculated numerically exactly
by carrying out the sums in Eqs.~\eqref{eq:RBM:Z} and~\eqref{eq:LossDKL}.
Similarly, the total correlations $\Ctot(p)$ of the target and model distributions are
computed exactly according to~\eqref{eq:Ctot}
as a sum over all states that keeps track of the contributions from both the full distribution $p(x)$ and the marginal ones $p_i(x_i)$.
 
For the partition function~\eqref{eq:RBM:Z},
we can exploit the bipartite structure of the RBM's interaction graph,
such that one of the sums can be factorized and thus be evaluated efficiently.
For example, if $\NH \leq \NV$,
we rewrite~\eqref{eq:RBM:Z} as
\begin{equation}
\label{eq:RBM:Z:Factorized}
	Z_\theta = \sum_h \e^{\sum_j b_j h_j} \prod_i \left( 1 + \e^{\sum_j w_{ij} h_j + a_i} \right) ,
\end{equation}
and similarly if $\NV < \NH$.
The sum over $h$ in~\eqref{eq:RBM:Z:Factorized} involves $2^{\NH}$ terms,
but the product over $i$ in each summand consists of just $\NV$ factors.
Therefore, the computational complexity scales exponentially with $\min\{ \NV, \NH \}$ only.
For the sum in Eq.~\eqref{eq:LossDKL},
we can exploit the sparsity of the target distribution $p(x)$ and restrict the (costly) evaluations of $\hat p_\theta(x)$ to those states with $p(x) > 0$.
Notwithstanding, the
system sizes for which the computation of $\Delta_\theta$ remains viable is relatively small;
see also Refs.~\cite{Desjardins:2010tmc, Fischer:2010ead, Schulz:2010icr, RomeroMerino:2018nbs, Manukian:2020mau}
for studies of the exact RBM loss in small examples.

In practical applications,
one does not have access to $p(x)$,
but only to a collection of samples $S := \{ \tilde x^{(1)}, \ldots, \tilde x^{(\lvert S \rvert)} \}$ (training and/or test data).
The empirical counterpart of $\Delta_\theta$ for such a dataset $S$ is
\begin{equation}
\label{eq:EmpiricalLossDKL}
\begin{aligned}
	\tilde\Delta_\theta^{(S)} 
	&= -\frac{1}{\lvert S \rvert} \sum_{x \in S} \left[ \sum_i a_i x_i + \sum_j \log\!\left( 1 + \e^{\sum_i w_{ij} x_i + b_j} \right) \right] \\
	& \qquad + \log Z_\theta -\log \lvert S \rvert \,;
\end{aligned}
\end{equation}
see also below Eq.~\eqref{eq:RBM:TrainingUpdate}.
The critical part is again the partition function $Z_\theta$.
Due to the aforementioned factorization (cf.\ Eq.~\eqref{eq:RBM:Z:Factorized}),
evaluating~\eqref{eq:EmpiricalLossDKL} remains feasible as long as the number of hidden units $\NH$ is sufficiently small,
even if $\NV$ is large.
Similarly, for small $\NH$, we can draw independent samples from $\hat p_\theta(x) = \sum_h \hat p_\theta(x \given h) \, \hat p_\theta(h)$,
without reverting to Markov chains and Gibbs sampling:
We first generate independent samples $\{ \tilde h^{(\mu)} \}$ of the hidden units,
using the fact that $\hat p_\theta(h)$ remains accessible for small $\NH$.
Subsequently, we sample configurations of the visible units using $\hat p_\theta(x \given h = \tilde h^{(\mu)})$.
This scheme was utilized to obtain the model test samples $\hat T$ for the $\NH \leq 32$ examples in Fig.~\ref{fig:Tradeoff:Applications}.
For the examples with $\NH > 32$, the samples in $\hat T$ were instead generated via Gibbs sampling according to~\eqref{eq:MarkovChainXH},
using $10$ parallel chains and storing every $\tau_\theta$-th sample after $2\times 10^6$ burn-in steps.

The accuracy measures $\tilde\delta_\theta$ and $\tilde\ell^1_\theta$ involve empirical distributions of coarse-grained visible-unit samples.
These reduced samples are obtained by using a weighted majority rule:
For a partition $\{ L_1, \ldots, L_{L} \}$ of the visible-unit indices $\{ 1, \ldots, \NV \}$ and a threshold $r \in [0,1]$,
we define
\begin{equation}
\label{eq:Reduction}
	f_\alpha(x) := \begin{cases}
		1 \text{ if } \sum_{i \in L_\alpha} x_i \geq r \, \lvert L_\alpha \rvert; \\
		0 \text{ otherwise}.
	\end{cases}
\end{equation}
For every sample $\tilde x$ in a given multiset $S$,
the associated coarse-grained sample is $\tilde y = (\tilde y_1, \ldots, \tilde y_L)$ with $\tilde y_\alpha := f_\alpha(\tilde x)$.

\subsection*{Details on $\tau_\theta$}

To measure the efficiency of Gibbs sampling according to the Markov chain~\eqref{eq:MarkovChainXH},
we evaluate the integrated autocorrelation time $\tau_\theta$ from~\eqref{eq:iact}.
The general purpose of Gibbs sampling is to estimate the model average $\< f(x) \> \equiv \< f(x) \>_{\hat p_\theta(x)}$
of some observable $f(x)$, i.e., a function of the visible units.
The sample mean $\bar f := \frac{1}{R} \sum_{n=0}^{R-1} f(x^{(n)})$ over a chain of $R$ samples
is an unbiased estimator of $\< f(x) \>$
if the chain is initialized and thus remains in the stationary regime, $x^{(0)} \sim \hat p_\theta(x)$ (see also below Eq.~\eqref{eq:iact}).
The correlation function associated with $f(x)$ and the Markov chain~\eqref{eq:MarkovChainXH} is
\begin{equation}
\label{eq:cf:General}
	g_\theta^{(f)}(n) := \< f(x^{(0)}) f(x^{(n)}) \> - \< f(x) \>^2 \,.
\end{equation}
For any such correlation function $g_\theta^{(f)}(n)$,
the corresponding integrated autocorrelation time is defined similarly to Eq.~\eqref{eq:iact},
\begin{equation}
\label{eq:iact:General}
	\tau_\theta^{(f)} := 1 + 2 \sum_{n=1}^\infty \frac{g^{(f)}_\theta(n)}{g^{(f)}_\theta(0)} \,.
\end{equation}
To assess the reliability of the estimator $\bar f$,
we inspect its variance
\begin{equation}
\label{eq:iact:ObsVar}
	\< \bar f^2 \> - \< \bar f \>^2 = \frac{g_\theta^{(f)}(0)}{R} \left[ 1 + 2 \sum_{n=1}^{R-1} \left(1 - \frac{n}{R} \right) \frac{g_\theta^{(f)}(n)}{g_\theta^{(f)}(0)} \right] .
\end{equation}
If the number of samples $R$ is much larger than the decay scale of $g_\theta^{(f)}(n)$ with $n$
(which is a prerequisite for estimating $\bar f$ reliably),
the contribution proportional to $\frac{n}{R}$ becomes negligible in the sum and
the term in brackets reduces to $\tau_\theta^{(f)}$ from~\eqref{eq:iact:General};
see also Sec.~2 of Ref.~\cite{Sokal:1997mcm}.
Observing that $g_\theta^{(f)}(0)$ is the variance of $f(x)$,
the variance of the estimator $\bar f$ from correlated Markov-chain samples is thus a factor of $\tau_\theta^{(f)}$ larger than the variance of the mean over independent samples.
In other words, sampling via the Markov chain~\eqref{eq:MarkovChainXH} requires $\tau_\theta^{(f)}$ more samples than independent sampling to reach the same standard error and is thus less efficient the larger $\tau_\theta$ becomes.

In general,
the integrated autocorrelation times $\tau_\theta^{(f)}$ can and will be different for different observables $f(x)$.
The specific choice $\tau_\theta$ from~\eqref{eq:iact}
is supposed to capture the generic behavior of typical observables.
It focuses on the individual visible units $x_i$ as the elementary building blocks.
However, instead of taking the mean over the autocorrelation times $\tau_\theta^{(x_i)}$ for each unit $f(x) = x_i$,
the averaging is performed at the level of the correlation functions $g_\theta^{(x_i)}(n)$;
cf.\ below Eq.~\eqref{eq:iact}.
The effect is a weighted average
\begin{equation}
\label{eq:iactAsVisUnitAvg}
	\tau_\theta = \frac{\sum_i g_\theta^{(x_i)}(0) \, \tau_\theta^{(x_i)} }{ \sum_i g_\theta^{(x_i)}(0) }
\end{equation}
that gives higher importance
to strongly fluctuating units with a large variance $g_\theta^{(x_i)}(0)$.
This accounts for the fact that variability of the Markov-chain samples is more important for those units and reduces the risk of underestimating correlations when there are certain regions in the data that behave essentially deterministically,
e.g., background pixels at the boundary of an image distribution.

In practice, if one is interested in a specific observable $f(x)$,
the associated autocorrelation time $\tau_\theta^{(f)}$ should be monitored directly instead of (or along with) the generic $\tau_\theta$.
While the quantitative details may differ,
we expect that the scaling behavior and the tradeoff mechanism remain qualitatively the same.
A comparison for different observables in the TFIC example from Fig.~\ref{fig:Tradeoff:TFIM} and in the digit-pattern images from Fig.~\ref{fig:Tradeoff:Applications}a--c can be found in Supplementary Note~4.
We indeed observe that $\tau_\theta^{(f)}$ is usually largely proportional to $\tau_\theta$.

In our numerical experiments, we estimate
$\tau_\theta$
statistically from long Markov chains of the form~\eqref{eq:MarkovChainXH} with $n_{\mathrm{tot}}$ samples.
Due to sampling noise,
the sum over time lags $n$ in~\eqref{eq:iact} must be truncated at a properly chosen threshold $n_{\max}$
to balance the bias and variance of the estimator.
Following Ref.~\cite{Sokal:1997mcm},
we choose $n_{\max}$ as the smallest integer such that $n_{\max} \geq \gamma \, \tilde\tau_\theta(n_{\max})$,
where $\gamma$ is a constant
and $\tilde\tau_\theta(n_{\max})$ is the value obtained from truncating~\eqref{eq:iact} at $n_{\max}$
using empirical averages to estimate the correlation function $g_\theta(n)$ (see below Eq.~\eqref{eq:iact})
and exploiting translational invariance of the stationary state (i.e., $\< x_i^{(0)} x_i^{(n)} \> = \< x_i^{(k)} x_i^{(n + k)} \>$).
If $g_\theta(n)$ follows an exponential decay,
the bias of the estimator is of order $\e^{-\gamma}$,
and we use $\gamma = 5$ in Figs.~\ref{fig:Tradeoff:TFIM}--\ref{fig:Tradeoff:PH1} and $\gamma = 8$ in Fig.~\ref{fig:Tradeoff:Applications}.
To reach the stationary regime, we initialize the chain~\eqref{eq:MarkovChainXH} in a state sampled uniformly at random and thermalize it by discarding a large number of samples,
at least on the order of $100 \tau_\theta$,
providing a reasonable buffer to account for mixing times that may exceed $\tau_\theta$
(and would thus increase the bias if the number of discarded samples was too small).

In Fig.~\ref{fig:Tradeoff:Applications},
we additionally maintain $r_{g}$ independently initialized chains to estimate $g_\theta(n)$
and calculate $\tau_\theta$ as described above,
using the average over the $r_g$ chains for $g_\theta(n)$.
The estimates are considered to be reliable only if the variations between the means of the $r_g$ chains are below $5 \,\%$;
otherwise the data points are discarded.
Furthermore, we repeat the entire procedure $r_\tau$ times,
leading to $r_\tau$ independent estimates of $\tau_\theta$.
The error bars in Fig.~\ref{fig:Tradeoff:Applications} indicate the min-max spread between those $r_\tau$ estimates.

\subsection*{Power-law bound}

In the examples from Figs.~\ref{fig:Tradeoff:TFIM}--\ref{fig:Tradeoff:Applications},
the blue dashed lines indicate the power-law bound~\eqref{eq:Tradeoff} for the accuracy--efficiency tradeoff.
The constants $c$ and $\alpha$ in this bound as stated in the respective figure panels were determined as follows:
The exponent $\alpha$ is chosen to roughly match the average slope $-\frac{\partial \log\Delta_\theta}{\partial\log\tau_\theta}$ for the data points in the correlation-learning regime over all hyperparameter configurations ($\nCD$, $\eta$, $B$, $\lvert S \rvert$) for any specific target distribution $p(x)$.
If this choice is ambiguous (e.g., in Fig.~\ref{fig:Tradeoff:TFIM}d),
the behavior in the beginning of the correlation-learning regime ($\tau_\theta \simeq 1$, $\Delta_\theta \simeq \Ctot(p)$) is decisive.
Once $\alpha$ is fixed,
$c$ is chosen as the maximum value such that $\Delta_\theta \tau_\theta^{\,\alpha} \geq c$ holds for all data points of all hyperparameter configurations simultaneously.

\subsection*{Examples}

The first examplary task (cf.\ Fig.~\ref{fig:Tradeoff:TFIM}) is quantum-state tomography,
namely to learn the ground-state wave function of the transverse-field Ising chain (TFIC)
based on measurements of the magnetization in a fixed spin basis $\{ \ket{x_1 \cdots x_M} \}$, where $x_i = 0$ ($x_i = 1$) indicates that the $i$th spin points in the ``up'' (``down'') direction in the chosen basis.
The Hamiltonian is $H = -\frac{1}{2} \sum_{i=1}^{\NV} (\sigma^x_i \sigma^x_{i+1} + g \, \sigma^z_i)$ with periodic boundary conditions and Pauli matrices $\sigma^{\gamma}_i$ ($\gamma = x,y,z$) acting on site $i$.
The model exhibits a quantum critical point at $\lvert g \rvert = 1$ and is integrable,
such that the ground state $\ket\psi = \sum_x \psi(x) \ket{x_1 \cdots x_M}$ can be constructed explicitly \cite{Pfeuty:1970odi, Vidmar:2016gge} (see also Supplementary Note~2A).
As we consider measurements in the $\sigma^z$ and $\sigma^x$ directions only,
the basis states $\ket{x_1 \cdots x_M}$ can be chosen such that
$\psi(x)$ is real-valued and nonnegative,
which allows us to employ the standard RBM architecture~\eqref{eq:RBM:pHX}.
(Generalizations for complex-valued wave function are possible \cite{Torlai:2018nnq, Czischek:2019ssn}.)
The target distribution is thus $p(x) = \psi(x)^2$.

Our second example (cf.\ Fig.~\ref{fig:Tradeoff:PH1}) is closer in spirit to traditional machine-learning applications
and involves pattern recognition and artificial image generation.
The target distribution $p(x)$ generates $5 \times 5$ pixel images
with a ``hook'' pattern comprised of $15$ pixels (see Fig.~\ref{fig:Tradeoff:PH1}a) implanted at a random position in a background of noisy pixels that are independently activated (white, $x_i = 1$) with probability $q = 0.1$ (see also Supplementary Note~2B for more details).
Periodic boundary conditions are assumed, meaning that $p(x)$ is translationally invariant along the two image dimensions.

We also consider a one-dimensional variant of this example with only $\NV = 4$ ($\NV = 5$) visible units
and an implanted ``010'' (``0110'') pattern, cf.\ Fig.~\ref{fig:Tradeoff:PH1}d.
In this case, we can solve the continuous-time learning dynamics ($\eta \to 0$ limit of~\eqref{eq:RBM:TrainingUpdate}) for the exact target and model distributions $p(x)$ and $\hat p_\theta(x, h)$,
obviating
artifacts caused by insufficient training data or biased gradient approximations,
see also Supplementary Note~1.

Our third example (cf.\ Fig.~\ref{fig:Tradeoff:Applications}a--c) is a simplified digit reproduction task.
Patterns of the ten digits $0$ through $9$ (see Fig.~\ref{fig:Tradeoff:Applications}a) are selected and inserted uniformly at random into image frames of $5 \times 7$ pixels,
with the remaining pixels outside of the pattern again activated with probability $q = 0.1$ (see Supplementary Note~2C for details).
No periodic boundary conditions are imposed, i.e., the input comprises proper, ordinary images.

In our fourth example (cf.\ Fig.~\ref{fig:Tradeoff:Applications}d,e),
we train RBMs on the MNIST dataset \cite{LeCun:YYYYmdh},
which consists of $28\times 28$-pixel grayscale images of handwritten digits.
It comprises a training set of $60\,000$ and a test set of $10\,000$ images.
We convert the grayscale images with pixel values between $0$ and $255$ to binary data by mapping values $0 \ldots 127$ to $0$ and $128 \ldots 255$ to $1$ (see also Supplementary Note~2D).


\section*{Acknowledgments}

This work was supported by KAKENHI Grant No.~JP22H01152 from the Japan Society for Promotion of Science.

The computer code for the numerical experiments can be accessed from the public repository \url{https://gitlab.com/lennartdw/xminirbm}.


%

\vspace{10pt}


\clearpage

\onecolumngrid
\begin{center}
\ \\[10pt]
{\large\textbf{SUPPLEMENTARY INFORMATION}}
\end{center}
\vspace{25pt}
\twocolumngrid

\setcounter{section}{0}
\renewcommand{\thesection}{S\arabic{section}}
\setcounter{figure}{0}
\renewcommand{\thefigure}{S\arabic{figure}}
\setcounter{equation}{0}
\renewcommand{\theequation}{S\arabic{equation}}
\setcounter{table}{0}
\renewcommand{\thetable}{S\arabic{table}}

Labels of equations, figures, and tables in these Supplementary Notes are prefixed by a capital letter ``S'' (e.g., Fig.~S1, Eq.~(S3)).
Any plain labels (e.g., Fig.~1, Eq.~(3), Ref.~[2]) refer to the corresponding items in the main text.

\section{Training details}
\label{ssec:Model:Training}

With the exception of Fig.~\mref{fig:Tradeoff:PH1}e,
the data presented in the main text
were obtained from RBMs trained with
a stochastic gradient descent scheme
based on the ideal gradient descent updates from Eq.~\meqref{eq:RBM:TrainingUpdate} of the main text
and utilizing contrastive divergence (CD, see Refs.~\cite{Hinton:2002tpe, Hinton:2012pgt}) or persistent contrastive divergence (PCD, see Ref.~\cite{Tieleman:2008trb}) to approximate the model averages.
Concretely,
the training dataset $S = \{ \tilde x^{(1)}, \ldots, \tilde x^{(\lvert S \rvert)} \}$
was partitioned randomly into $s := \frac{\lvert S \rvert}{B}$ minibatches $S_1, \ldots, S_s$ of size $B$ at the beginning of each epoch $t$.
We recall that Markov chains of the form
\begin{equation}
\label{eq:S:MarkovChainXH}
	x^{(0)} \rightarrow h^{(0)} \rightarrow x^{(1)} \rightarrow h^{(1)} \rightarrow \cdots
\end{equation}
are employed to assess the model distribution approximately (cf.\ Eq.~\meqref{eq:MarkovChainXH} of the main text).
We denote a particular (random) realization of $x^{(n)}$ and $h^{(n)}$ for a chain initiated at a (fixed) $x^{(0)} = \tilde x$ by $\hat x^{(n)}(\tilde x)$ and $\hat h^{(n)}(\tilde x)$, respectively.
In CD,
the updates~\meqref{eq:RBM:TrainingUpdate} are then approximated as
\begin{subequations}
\label{eq:S:RBM:TrainingUpdate:CD}
\begin{align}
	& w_{ij}(t + \tfrac{r}{s}) - w_{ij}(t + \tfrac{r-1}{s}) \notag \\
		&\;= \frac{\eta}{B} \sum_{\tilde x \in S_r} \left[ \tilde x_i \hat h_j^{(0)}(\tilde x)  - \hat x_i^{(\nCD)}(\tilde x) \hat h_j^{(\nCD)}(\tilde x) \right] , \\
	& a_{i}(t + \tfrac{r}{s}) - a_{i}(t + \tfrac{r-1}{s}) \notag \\
		&\;= \frac{\eta}{B} \sum_{\tilde x \in S_r} \left[ \tilde x_i - \hat x_i^{(\nCD)}(\tilde x) \right] , \\
	& b_{j}(t + \tfrac{r}{s}) - b_{j}(t + \tfrac{r-1}{s}) \notag \\
		&\;= \frac{\eta}{B} \sum_{\tilde x \in S_r} \left[ \hat h_j^{(0)}(\tilde x)  - \hat h_j^{(\nCD)}(\tilde x) \right]
\end{align}
\end{subequations}
for $r = 1, \ldots, s$.
In essence,
the model averages in~\meqref{eq:RBM:TrainingUpdate} are thus approximated by empirical averages over samples from Markov chains~\eqref{eq:S:MarkovChainXH}
that are initialized with a training sample $x^{(0)} = \tilde x \in S$.
If $\Delta_\theta$ is sufficiently small, such $x^{(0)}$ from $S$ may already be a reasonable approximation for a sample from $\hat p_\theta(x)$,
and the chain generates a new (but correlated) sample.
In the beginning of training,
when $\hat p_\theta(x)$ is still far from $p(x)$,
such an initialization of the chains is less justified,
but it is found to work in practice \cite{Hinton:2012pgt, Fischer:2012irb, Montufar:2018rbm},
not least because the mixing and autocorrelation times of the chains are typically small as well in this case.

In PCD,
the updates~\meqref{eq:RBM:TrainingUpdate} are approximated as
\begin{subequations}
\label{eq:S:RBM:TrainingUpdate:PCD}
\begin{align}
	& w_{ij}(t + \tfrac{r}{s}) - w_{ij}(t + \tfrac{r-1}{s}) \notag \\
		&\;= \eta \left[  \frac{1}{B} \sum_{\tilde x \in S_r} \tilde x_i \hat h_j^{(0)}(\tilde x) - \frac{1}{L} \!\!\!\!\! \!\! \sum_{x' \in Q(t + \frac{r}{s})} \!\!\!\!\! \!\! x'_i \hat h_j^{(0)}(x') \right] , \\
	& a_{i}(t + \tfrac{r}{s}) - a_{i}(t + \tfrac{r-1}{s}) \notag \\
		&\;= \eta \left[ \frac{1}{B} \sum_{\tilde x \in S_r} \tilde x_i - \frac{1}{L} \!\!\!\!\! \!\! \sum_{x' \in Q(t + \frac{r}{s})} \!\!\!\!\! \!\! x'_i \right] , \\
	& b_{j}(t + \tfrac{r}{s}) - b_{j}(t + \tfrac{r-1}{s}) \notag \\
		&\;= \eta \left[ \frac{1}{B} \sum_{\tilde x \in S_r} \hat h_j^{(0)}(\tilde x)  - \frac{1}{L} \!\!\!\!\! \!\! \sum_{x' \in Q(t + \frac{r}{s})} \!\!\!\!\! \!\! \hat h_j^{(0)}(x') \right] , \\
	& Q(t + \tfrac{r}{s}) 
		= \left\{ \hat x^{(\nCD)}(x') : x' \in Q(t + \tfrac{r-1}{s}) \right\}
\end{align}
\end{subequations}
for $r = 1, \ldots, s$, where $Q(0)$ is a set of random, independent configurations of the visible units and $L := \lvert Q(0) \rvert$.
We always use $L = B$.
Hence the model averages are approximated by empirical averages over samples from Markov chains~\eqref{eq:S:MarkovChainXH} that are initialized with (or, in other words, continued from) samples of the previous update step,
i.e., the chains are \emph{persistent}.
Note, however, that the model distribution used to generate (or advance) the chains changes from step to step as the model parameters $\theta$ change.
If these parameter updates are sufficiently small and the initial configurations $x' \in Q(0)$ emulate samples from the initial model $\hat p_{\theta(0)}(x)$,
the chains approximately reflect the gradually evolving stationary model distribution throughout the training process.
However, subsequent samples are generally not independent,
especially if $\nCD \lesssim \tau_\theta$.

The total number of training epochs in Figs.~\mref{fig:Tradeoff:TFIM}--\mref{fig:Tradeoff:Applications} of the main paper
as well as in this Supplementary Information
varies between $2 \times 10^4$ and $3 \times 10^6$,
depending on the time needed until improvement of the accuracy could no longer be observed
and extending reasonably far beyond it to capture the degradation regime.
As mentioned in the main text,
the
machines
were initialized by drawing the parameters $\theta_k(0)$ independently from a normal distribution $\mc N(\mu, \sigma)$ of mean $\mu$ and standard deviation $\sigma$, namely
$w_{ij}(0) \sim \mc N(0, 10^{-2})$ and
$a_i(0), b_j(0) \sim \mc N(0, 10^{-1})$ 
(Figs.~\mref{fig:Tradeoff:TFIM}, \mref{fig:Tradeoff:PH1}, and~\mref{fig:Tradeoff:Applications}b,c)
or $a_i = b_j = 0$ (Fig.~\mref{fig:Tradeoff:Applications}e).
Figs.~\mref{fig:Tradeoff:TFIM} and~\mref{fig:Tradeoff:PH1} show averages over
$5$ independent repetitions of the experiment for each hyperparameter configuration.
Fig.~\mref{fig:Tradeoff:Applications} shows results for single machines.

For the data in Fig.~\mref{fig:Tradeoff:PH1}e,
we utilized---as described in the main text---the full target distribution $p(x)$ and the exact $n$-step CD model distribution $\hat p^{(n)}_\theta$ or the full model distribution $\hat p_\theta(x)$ for the expectation values.
Moreover, we employed
the continuous-time limit of the update equations~\meqref{eq:RBM:TrainingUpdate}.
Hence the evolution of the weights is governed by the differential equations
\begin{subequations}
\label{eq:S:RBM:TrainingUpdateExact}
\begin{align}
	\dot w_{ij}(t)
		&= \sum_{x, h} x_i h_j \, \hat p_{\theta(t)}(h \given x) \left[  p(x) - \hat p^{(n)}_{\theta(t)}(x) \right] ,\\
	\dot a_{i}(t)
		&= \sum_{x} x_i \left[ p(x) - \hat p^{(n)}_{\theta(t)}(x) \right] , \\
	\dot b_{j}(t)
		&= \sum_{x, h} h_j \, \hat p_{\theta(t)}(h \given x) \left[  p(x) - \hat p^{(n)}_{\theta(t)}(x) \right] ,
\end{align}
\end{subequations}
where the dots indicate derivatives with respect to $t$ and
\begin{align}
	\hat p_\theta^{(n)}(x)
		&:= \sum_{x', h'} \hat p_\theta(x \given h') \, \hat p_\theta(h' \given x') \hat p_\theta^{(n-1)}(x') \,, \\
	\hat p_\theta^{(0)}(x) &:= p(x) \,,
\end{align}
and $\hat p_\theta^{(\infty)}(x) \equiv \hat p_\theta(x)$.
We then integrated Eqs.~\eqref{eq:S:RBM:TrainingUpdateExact} numerically (starting from random initial conditions as before)
using \emph{Mathematica}'s \texttt{NDSolve} routine.

\section{Example tasks}
\label{ssec:Examples}

\subsection{Transverse-field Ising chain}
\label{ssec:Examples:TFIM}

The transverse-field Ising chain (TFIC) with $M$ sites is defined by the Hamiltonian
\begin{equation}
\label{eq:S:TFIM:H}
	H = -\frac{1}{2} \sum_{i = 0}^{M-1} \left( \sigma_i^x \sigma_{i+1}^x + g \, \sigma_i^z \right)
\end{equation}
with periodic boundary conditions, $\sigma_{i+M}^\gamma = \sigma_i^\gamma$,
where $\sigma_i^\gamma$ ($\gamma = x, y, z$) are the Pauli matrices acting on site $i$.
The corresponding spin raising and lowering operators are $\sigma_i^\pm := \frac{1}{2} (\sigma_i^x \pm \I \sigma_i^y)$.

The Hamiltonian can be diagonalized by the following sequence of transformations:
the Jordan-Wigner transformation
\begin{equation}
	c_i := P_i \sigma^-_i
\end{equation}
with $P_i := \prod_{j=0}^{i-1} (-\sigma_i^z)$,
the Fourier transformation
\begin{equation}
\tilde c_k := \frac{1}{\sqrt{L}} \sum_i \e^{-\I k i} c_i
\end{equation}
with fermionic (bosonic) Matsubara frequencies $k$ if the total particle number $\sum_i c_i^\dagger c_i$ is even (odd),
and the Bogoliubov transformation
\begin{equation}
	\eta_k := u_k \tilde c_k - v_k \tilde c_{-k}^\dagger
\end{equation}
with $u_k := (\varepsilon_k + \alpha_k)/\omega_k$, $v_k := \I \beta_k / \omega_k$, $\alpha_k := -2 J (g + \cos k)$, $\beta_k := 2 J \, \sin k$, $\varepsilon_k^2 := \alpha_k^2 + \beta_k^2$, $\omega_k^2 := 2 \varepsilon_k (\varepsilon_k + \alpha_k)$;
see, for example, Ref.~\cite{Vidmar:2016gge}.
The resulting Hamiltonian is
\begin{equation}
\label{eq:S:TFIM:HDiag}
	H = \sum_k \varepsilon_k \left( \eta_k^\dagger \eta_k - \tfrac{1}{2} \right)
\end{equation}
in the sector with an even number of particles,
to which the ground state belongs.
This ground state can be constructed as
\begin{equation}
\label{eq:S:TFIM:GroundState}
	\ket\psi := \prod_k \frac{1}{\lvert v_k \rvert} \eta_k \eta_{-k} \ket{ \downarrow \cdots \downarrow }
\end{equation}
from the state $\ket{ \downarrow \cdots \downarrow }$ with all spins down in the $\sigma^z$ basis \cite{Vidmar:2016gge}.

Moreover, by adjusting the global phase,
it can be written such that $\psi(x) := \< x_0 \cdots x_{M-1} | \psi \>$ is real-valued and nonnegative when $\ket{ x_0 \cdots x_{M-1} }$ is a basis state in the $\sigma^z$ or $\sigma^x$ bases.
For our quantum-state tomography task of the ground-state wave function from measurements in either of the two bases,
we can therefore take the target distribution as $p(x) := \psi(x)^2$ and do not need additional modifications of the RBM model to facilitate the phase reconstruction \cite{Torlai:2018nnq}.

Tab.~\ref{tab:TFIMCharacteristics} lists the entropy $S(p) := -\sum_x p(x) \, \ln p(x)$ and total correlation $\Ctot(p)$ (cf.\ Eq.~\meqref{eq:Ctot} of the main text) of the target distribution obtained for various values of $g$.

\begin{table}
\caption{Entropy $S(p)$ and total correlation $\Ctot(p)$ of the TFIC ground-state distribution in the $\sigma^z$ and $\sigma^x$ bases for various values of the magnetic-field strength $g$.}
\label{tab:TFIMCharacteristics}
\centering
\begin{tabularx}{\linewidth}{c c c Z Z c Z Z c}
\hline\hline
& & & \multicolumn{2}{c}{$\sigma^z$ basis} & & \multicolumn{2}{c}{$\sigma^x$ basis} &  \\
\hspace{0.05\linewidth} & \ $g$\  & \hspace{0.1\linewidth} & \mbox{$S(p)$} & \mbox{$\Ctot(p)$} & \hspace{0.1\linewidth} & \mbox{$S(p)$} & \mbox{$\Ctot(p)$} & \hspace{0.05\linewidth} \\
\hline
& 0.5 & & 12.48 & 0.705 & & 1.216 & 12.65 & \\
& 0.8 & & 10.99 & 0.824 & & 5.062 & 8.801 & \\
& 1   & & 8.028 & 1.441 & & 8.721 & 5.142 & \\
& 1.2 & & 4.808 & 1.891 & & 11.38 & 2.478 & \\
& 2   & & 1.760 & 1.134 & & 13.17 & 0.689 & \\
& 4   & & 0.523 & 0.400 & & 13.70 & 0.160 & \\
\hline\hline
\end{tabularx}
\end{table}

\subsection{Hook-pattern images}
\label{ssec:Examples:PH1}

We consider $L\times L$ images $x = (x_{i_1,i_2})_{i_1,i_2=0}^{L-1}$
whose pixels $x_i = x_{i_1,i_2}$ are either black $(x_{i} = 0)$ or white ($x_{i} = 1$).
We denote the set of all $2^{L \times L}$ of these images by $\mc K$.

The characteristic feature of the target distribution $p(x)$ is a ``hook'' pattern comprised of a total of $15$ pixels, cf.\ Fig.~\mref{fig:Tradeoff:PH1}a.
We assume periodic boundary conditions ($x_{i,j} = x_{i+L,j} = x_{i,j+L}$)
and denote the $\max$-distance on the image grid by
\begin{equation}
	d_\infty(i, j) = \max_{\mu = 1,2} \min\{ \lvert i_\mu - j_\mu \rvert, L - \lvert i_\mu - j_\mu \rvert \} \,.
\end{equation}
Similarly, for a set of pixel positions (sites) $I$ and a single site $j$,
define $d_\infty(I, j) := \min\{ d_\infty(i, j) : i \in I \}$.
An image $x$ then shows the ``hook'' pattern at site $i$
if the pixels at sites $I_0(i) := \{ (i_1, i_2), (i_1 - 1, i_2), (i_1, i_2 + 1) \}$ are white
\emph{and} the pixels at sites $I_1(i) := \{ j : d_\infty(I_0(i), j) = 1 \}$ are black.
In other words,
$x_j = 1$ for all $j \in I_0(i)$ and $x_j = 0$ for all $j \in I_1(i)$.
The remaining $L^2 - 15$ pixels at sites $I_{\geq 2}(i) := \{ j : d_\infty(I_0(i), j) \geq 2 \}$
can take arbitrary values.
We denote the set of all images with a ``hook'' pattern at an arbitrary site by
$\KPH$.
In the main text, we choose $L = 5$;
examples are shown in Fig.~\mref{fig:Tradeoff:PH1}a.
Note that there are a total of $\lvert \KPH \rvert = L^2 \times 2^{L^2 - 15} = 25\,600$ images exhibiting the pattern,
which is a fraction of $\lvert \KPH \rvert / \lvert \mc K \rvert = 25 \times 2^{-15} \approx 0.08\,\%$ of all $5\times 5$ images.

The target distribution picks a random location $i = (i_1, i_2)$ for the hook pattern and activates all remaining pixels in $I_{\geq 2}(i)$ with probability $q = \frac{1}{10}$.
Hence
\begin{equation}
\label{eq:S:PH1:pX}
	p(x) = \id_{\KPH}(x) \sum_i \prod_{j \in I_{\geq 2}(i)} \left[ x_j \, q + (1 - x_j) (1 - q) \right] ,
\end{equation}
where $\id_S$ is the indicator function of the set $S$,
i.e., $\id_S(x) = 1$ if $x \in S$ and $0$ otherwise.

The entropy is $S(p) \approx 6.47$
and the total correlation is $\Ctot(p) \approx 4.53$.

The one-dimensional, smaller distributions from Fig.~\mref{fig:Tradeoff:PH1}d
are structurally similar,
but involve a core pattern of white pixels at sites $I_0(i)$ of only one ($M = 4$) or two ($M = 5$) pixels,
surrounded by one-site boundaries $I_1(i)$ of black pixels in either direction,
with the remaining pixel in $I_{\geq 2}(i)$ being arbitrary again.

\subsection{Digit images}
\label{ssec:Examples:PN}

We consider $H \times W$ images $x = (x_{i_1,i_2})$ with pixel values $x_i = x_{i_1,i_2} \in \{ 0, 1 \}$
and denote the set of all such images by $\mc K$.

The target distribution involves images showing one of ten patterns representing the digits $0$ through $9$ (cf.\ Fig.~\mref{fig:Tradeoff:Applications}a of the main text).
The digit patterns consist of a core block of fixed black or white pixels (black or white in Fig.~\mref{fig:Tradeoff:Applications}a)
as well as a boundary (gray-shaded in Fig.~\mref{fig:Tradeoff:Applications}a)
which may either be represented by black pixels or by the image frame (i.e., the gray pixels may lie ``out of bounds'').
No periodic boundary conditions are assumed.
The remaining pixels outside of the respective pattern may take arbitrary values.
Due to the different sizes of the digit patterns,
the number of possible images for each pattern is different in general.
We denote the set of all images with a ``$k$'' pattern by $\mc K_k$.
For $H = 7$ and $W = 5$,
which is our choice for the data shown in Fig.~\mref{fig:Tradeoff:Applications}a--c of the main text,
the resulting cardinalities of the sets $\mc K_k$ are summarized in Tab.~\ref{tab:PNCounts}.

\begin{table}
\caption{Cardinalities of the sets $\mc K_k$ of digit images.
In total, $\lvert \mc K_{\mathrm{PN}} \rvert = 40\,507\,353$ out of the total of $2^{7 \times 5} = 34\,359\,738\,368$ possible images show digit patterns.}
\label{tab:PNCounts}
\centering
\begin{tabularx}{\linewidth}{c Z c c Z}
\hline\hline
\ $k$\  & $\lvert \mc K_k \rvert$ & \hspace{0.3\linewidth} & \ $k$\  & $\lvert \mc K_k \rvert$ \\
\hline
0 & $8\,513$ & & 5 & $565\,391$ \\
1 & $38\,558\,138$ & & 6 & $56\,464$ \\
2 & $565\,391$ & & 7 & $66\,624$ \\
3 & $565\,391$ & & 8 & $8\,513$ \\
4 & $56\,464$ & & 9 & $56\,464$ \\
\hline\hline
\end{tabularx}
\end{table}

The target distribution selects a pattern $k$ uniformly at random, $p(x \in \mc K_k) = \frac{1}{10} \id_{\bigcup_l \mc K_l}(x)$.
The selected pattern is then placed at a random position $i = (i_1, i_2)$ with uniform probability $p(x \in \mc K_{k,i} \given x \in \mc K_k)$,
where $\mc K_{k,i}$ is the set of all images showing pattern $k$ at position $i$.
Due to the different pattern shapes, again, the admissible positions are generally different for different patterns.
Denoting the set of all pixel sites that are not part of the pattern $k$ at position $i$ by $\bar I(k, i)$,
the remaining pixels represent noise and are white independently with probability $q = \frac{1}{10}$,
i.e.,
$p(x_j = 1 | j \in \bar I(i, k)) = q$.
The total probability of a given image $x$ is thus
\begin{equation}
\label{eq:S:PN:pX}
\begin{aligned}
	p(x) &= \sum_{k} p(x \in \mc K_k) \sum_i p(x \in \mc K_{k,i} \given x \in \mc K_k) \\
		& \quad \times \prod_{j \in I(k,i)} \left[ x_j \, q + (1 - x_j) (1 - q) \right] .
\end{aligned}
\end{equation}
Note that an image can ``accidentally'' have patterns at multiple positions;
for the image sizes we adopted,
in particular,
this can happen if one of them is the ``1'' pattern.

The entropy of this distribution is $S(p) \approx 8.35$ and the total correlation is $\Ctot(p) \approx 11.67$.

For the empirical loss measure $\tilde\Delta_\sigma^{(T, \hat T)}$ in Fig.~\mref{fig:Tradeoff:Applications}c,
the parameter $\sigma$ is determined by minimizing $\tilde\Delta_\sigma^{(T, S)}$ between the test and training datasets of $\lvert T \rvert = 10\,000$ and $\lvert S \rvert = 50\,000$ samples, respectively.
From the relationship shown in Fig.~\ref{fig:S:EmpGaussSigma}a,
we find $\tilde\Delta^{(T, S)}_\sigma \approx 1.354$ at $\sigma \approx 0.32$ for the minimum.

\subsection{MNIST}

We use the  MNIST dataset of $28\times 28$-pixel images of handwritten digits with its standard splitting into $\lvert S \rvert = 60\,000$ training and $\lvert T \rvert = 10\,000$ test images \cite{LeCun:YYYYmdh}.
Each subset contains equal fractions of representations for each digit.
The grayscale pixel values $z_i \in \{ 0, 1, \ldots, 255 \}$ are preprocessed as $x_i = \lfloor \frac{z_i}{128} \rfloor$ to obtain a binary dataset.
The entropies of the thus-obtained training and test datasets are
$S(\pEmp(\,\cdot\,; S)) \approx 11.00$ and $S(\pEmp(\,\cdot\,; T)) \approx 9.21$,
and their total correlations are
$\Ctot(\pEmp(\,\cdot\,; S)) \approx 195.0$ and $\Ctot(\pEmp(\,\cdot\,; T)) \approx 196.5$, respectively.

\begin{figure}
\centering
\includegraphics[width=\linewidth]{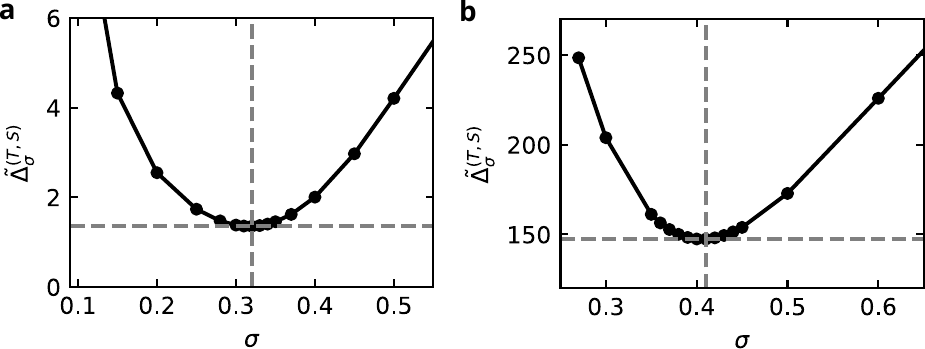}
\caption{Kullback-Leibler divergence $\tilde\Delta_\sigma^{(T, S)} := \DKL(\tilde p(\,\cdot\,; T) || \tilde p_\sigma(\,\cdot\,; S))$ between the empirical distribution of the test data $\tilde p(x; T)$ and the Gaussian-smoothened empirical distribution $\tilde p_\sigma(x; S)$ of the training data as a function of the smoothening width parameter $\sigma$.
The value of $\sigma$ for which $\tilde\Delta_\sigma^{(T, S)}$ becomes minimal is adopted in Fig.~\mref{fig:Tradeoff:Applications} to calculate the empirical loss measure $\tilde\Delta_\sigma^{(T, \hat T)}$.}
\label{fig:S:EmpGaussSigma}
\end{figure}

Similarly as before, the parameter $\sigma$ for $\tilde\Delta_\sigma^{(T, \hat T)}$ in Fig.~\mref{fig:Tradeoff:Applications}e is chosen such that $\tilde\Delta_\sigma^{(T, S)}$ becomes minimal.
As shown in Fig.~\ref{fig:S:EmpGaussSigma}b,
we obtain $\tilde\Delta^{(T, S)}_\sigma \approx 147.4$ at $\sigma \approx 0.41$ for that minimum.

\begin{figure*}
\centering
\includegraphics[scale=1]{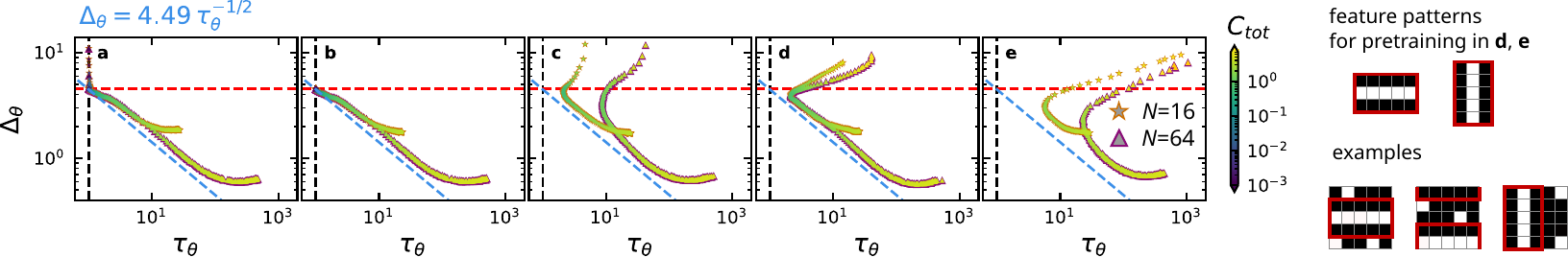}
\caption{Exact loss $\Delta_\theta$ vs.\ autocorrelation time $\tau_\theta$ for various initialization schemes of the weights in the pattern recognition task from Fig.~\mref{fig:Tradeoff:PH1} (see also Sec.~\ref{ssec:Examples:PH1}).
Initialization schemes:
(a) $w_{ij} \sim \mc N(0, 0.01)$, $a_i, b_j \sim \mc N(0, 0.1)$ (similar to the main text);
(b) $w_{ij} \sim \mc N(0, 0.01)$, $a_i = \ln[\nu_i / (1 - \nu_i)]$
with $\nu_i := \frac{1}{\lvert S \rvert} \sum_{\tilde x \in S} \tilde x_i$ the activation frequency of the $i$th visible unit in the training dataset $S$,
$b_j = 0$ \cite{Hinton:2012pgt};
(c) $w_{ij}, a_i, b_j \sim \mc N(0, 1)$;
(d) snapshot after $t = 500$ training epochs of a machine initialized like in (a) and subsequently trained on images with horizontal or vertical line patterns (see sketch) and otherwise identical hyperparameters;
(e) similar to (d), but using snapshots after $t = 50\,000$ training epochs.
Data points show averages over $5$ independent runs for each scheme.
Fill colors indicate the total correlation $\Ctot(\hat p_\theta)$ of the model distribution (see colorbar),
border colors and marker types indicate the number of hidden units $\NH$ (see legend in~(e)).
Further hyperparameters: CD training with $\nCD = 1$, $\eta = 0.005$, $B = 100$, $\lvert S \rvert = 5000$.
}
\label{fig:S:PH1:InitCond}
\end{figure*}

\begin{figure*}
\centering
\includegraphics[scale=1]{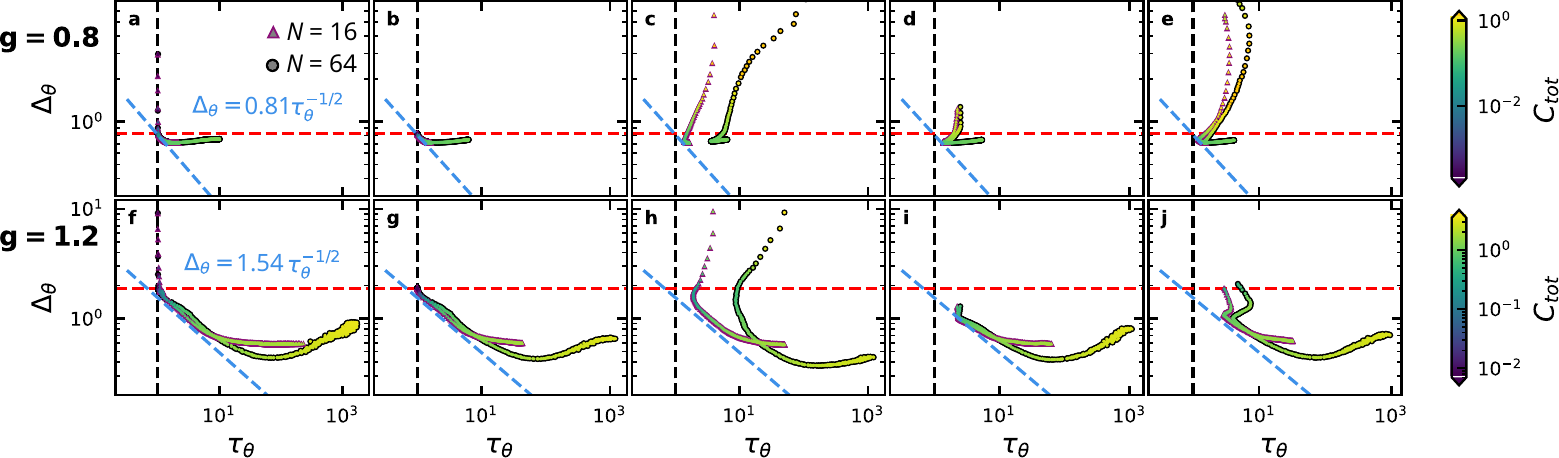}
\caption{Exact loss $\Delta_\theta$ vs.\ autocorrelation time $\tau_\theta$ for various initialization schemes of the weights in the ground-state tomography task for the TFIC with $g = 0.8$ (top row) and $g = 1.2$ (bottom row), cf.\ Fig.~\mref{fig:Tradeoff:TFIM} and Sec.~\ref{ssec:Examples:TFIM}).
Initialization schemes:
(a,f) $w_{ij} \sim \mc N(0, 0.01)$, $a_i, b_j \sim \mc N(0, 0.1)$ (similar to the main text);
(b,g) $w_{ij} \sim \mc N(0, 0.01)$, $a_i = \ln[\nu_i / (1 - \nu_i)]$
with $\nu_i := \frac{1}{\lvert S \rvert} \sum_{\tilde x \in S} \tilde x_i$ the activation frequency of the $i$th visible unit in the training dataset $S$,
$b_j = 0$ \cite{Hinton:2012pgt};
(c,h) $w_{ij}, a_i, b_j \sim \mc N(0, 1)$;
(d,i) snapshot after $t = 5000$ training epochs of a machine initialized like in (a,f) and subsequently trained to learn the ground state for $g = 1$ and otherwise identical hyperparameters;
(e,j) similar to (d,i), but using snapshots after $t = 5000$ training epochs when learning the $g = 2$ ground state.
Data points show averages over $5$ independent runs for each scheme.
Fill colors indicate the total correlation $\Ctot(\hat p_\theta)$ of the model distribution (see colorbars),
border colors and marker types indicate the number of hidden units $\NH$ (see legend in~(a)).
Further hyperparameters: CD training with $\nCD = 1$, $\eta = 0.001$, $B = 100$, $\lvert S \rvert = 25\,000$.
}
\label{fig:S:TFIM:InitCond}
\end{figure*}

\section{Initialization schemes}
\label{ssec:InitSchemes}

As argued in the main text,
the most natural way to initialize the RBM parameters $(\theta_k) = (w_{ij}, a_i, b_j)$ is to assign small random values to them
if no information about the target distribution $p(x)$ is available.
In the main text, we thus sample the initial $\theta_k$ from independent normal distributions $\mc N(\mu, \sigma)$ with vanishing mean $\mu = 0$ and small (or vanishing) standard deviation $\sigma$.
In Figs.~\ref{fig:S:PH1:InitCond} and~\ref{fig:S:TFIM:InitCond}, we compare the resulting relationship between $\Delta_\theta$ and $\tau_\theta$ for other initialization schemes.

One piece of information about the target distribution that is easily accessible
is the approximate value of the marginal probabilities $p_i(x_i)$ of the visible units.
Hence Hinton \cite{Hinton:2012pgt} suggests to initialize the visible-unit biases as $a_i = \ln[ \nu_i / (1 - \nu_i) ]$,
where $\nu_i := \frac{1}{\lvert S \rvert} \sum_{\tilde x \in S} \tilde x_i$ is the frequency of $x_i = 1$ in the training dataset $S$,
i.e., $p_i(x_i = 1) \approx \nu_i$.
In the numerical examples from the second columns of Figs.~\ref{fig:S:PH1:InitCond} and~\ref{fig:S:TFIM:InitCond}, we cap the so-obtained $a_i$ at $2$ in absolute value.
For the weights and hidden-unit biases, Ref.~\cite{Hinton:2012pgt} suggests using $w_{ij} \sim \mc N(0, 0.01)$ and $b_j = 0$.
Following this scheme,
one can shorten the independent-learning period as can be seen in the second columns of Figs.~\ref{fig:S:PH1:InitCond} and~\ref{fig:S:TFIM:InitCond}.
Nevertheless, we observe the same learning characteristics as for the fully random initialization
in the correlation-learning and degradation regimes.

By accident or deliberation,
the initial model distribution may already exhibit noticeable but spurious correlations as well.
This is examplified in the last three columns of Figs.~\ref{fig:S:PH1:InitCond} and~\ref{fig:S:TFIM:InitCond}.
Such initial correlations may arise, for instance,
if the weights are chosen too large (third columns) 
or if parameters from a pre-trained machine using a different target distribution are adopted (fourth and fifth columns).
In the case of such spurious initial correlations,
the machine typically starts further away from the lower bound~\meqref{eq:Tradeoff}.
As training progresses, however, the bound is approached by decreasing both $\Delta_\theta$ and $\tau_\theta$ first
and eventually showing similar tradeoff characteristics as in the ``independent'' initialization schemes (first two columns).

\begin{figure*}
\centering
\includegraphics[scale=1]{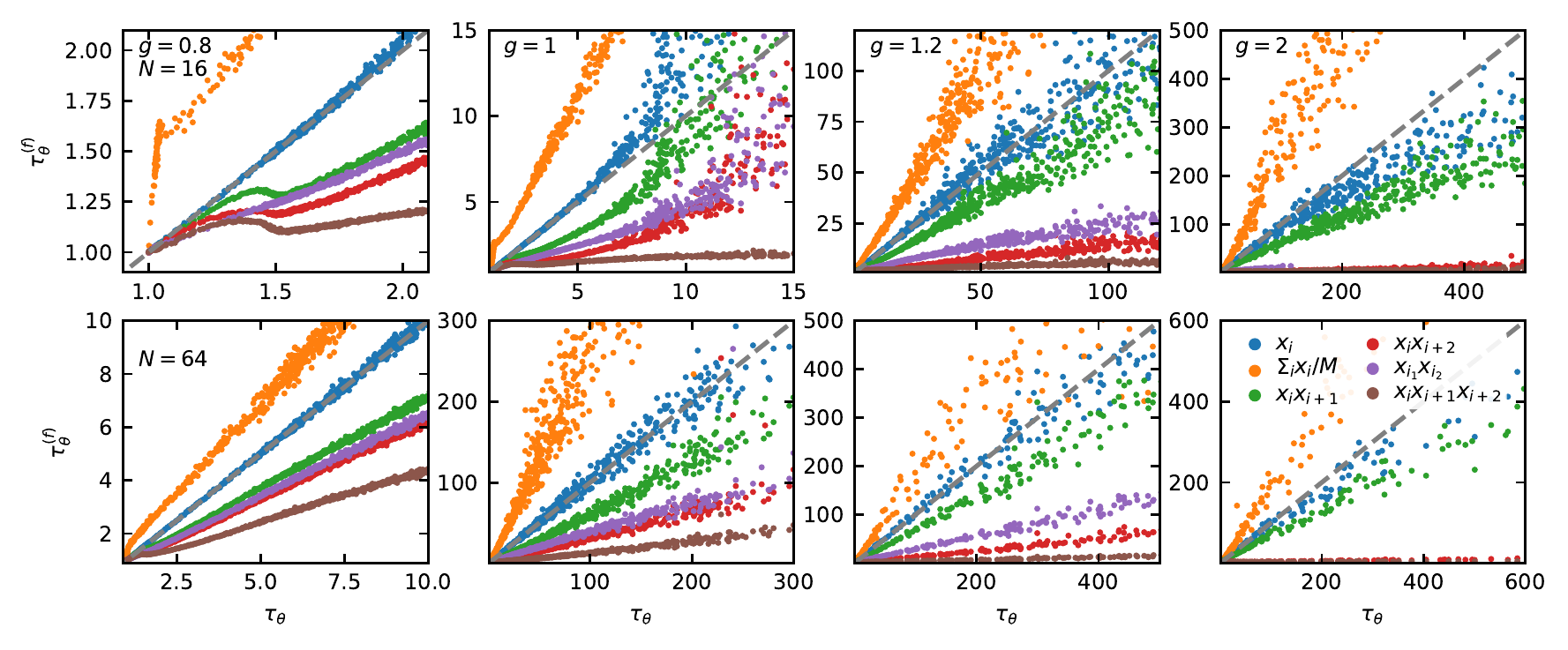}
\caption{Integrated autocorrelation times $\tau_\theta^{(f)}$ (cf.\ Eqs.~\eqref{eq:S:cf:2pointNN}--\eqref{eq:S:cf:mean}) for various observables $f(x)$ (see legend in the bottom-right panel) vs.\ sampling efficiency $\tau_\theta$ in the transverse-field Ising chain (TFIC, cf.\ Fig.~\mref{fig:Tradeoff:TFIM} and Sec.~\ref{ssec:Examples:TFIM}) in the $\sigma^z$ basis.
The panels show results for different external fields $g$ (cf.\ Eq.~\eqref{eq:S:TFIM:H}) in the columns
and different numbers of hidden units $\NH$ in the rows as indicated.
Hyperparameters: $\nCD = 1$, $\eta = 10^{-3}$, $B = 100$, $\lvert S \rvert = 25\,000$.
For observables depending on the visible-unit indices $i, i_1, i_2$,
the data are averaged over all those indices.
Furthermore, as in the main text, each data point is an average over five independently trained RBMs at a fixed training epoch $t$.
The gray dashed line shows the function $\tau_\theta^{(f)} = \tau_\theta$ for reference.}
\label{fig:S:TFIC:ZBasis:ObsIACTCompare}
\end{figure*}

\begin{figure*}
\centering
\includegraphics[scale=1]{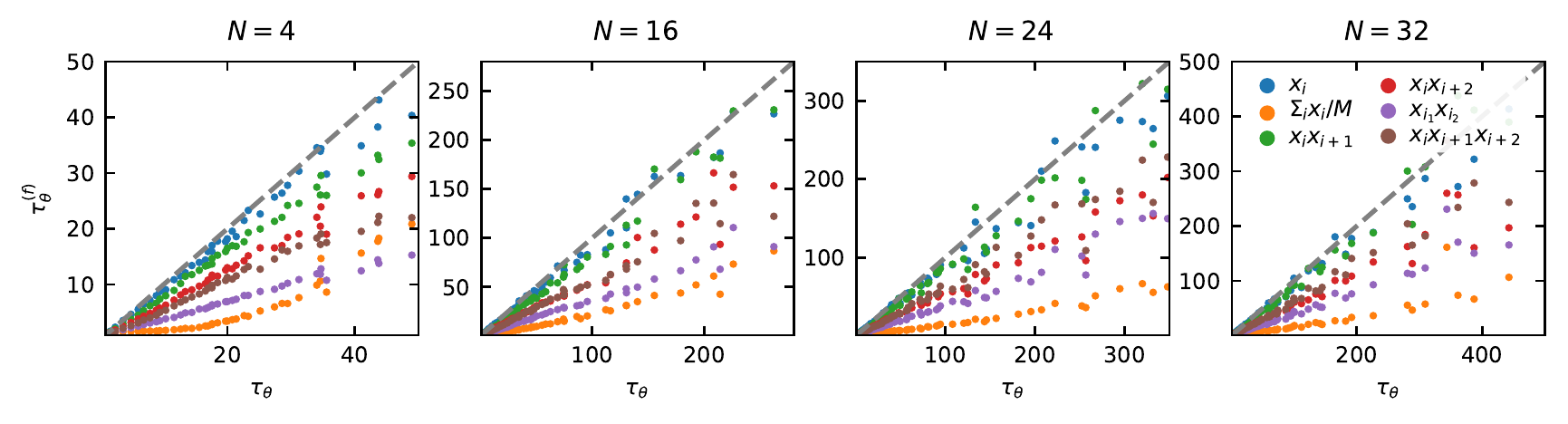}
\caption{Integrated autocorrelation times $\tau_\theta^{(f)}$ (cf.\ Eqs.~\eqref{eq:S:cf:2pointNN}--\eqref{eq:S:cf:mean}) for various observables $f(x)$ (see legend in the right-most panel) vs.\ sampling efficiency $\tau_\theta$ in the digit-pattern example from Fig.~\mref{fig:Tradeoff:Applications}a--c and Sec.~\ref{ssec:Examples:PN}.
The panels show results for different numbers of hidden units $\NH$ as indicated.
Hyperparameters: $\nCD = 1$, $\eta = 0.005$, $B = 100$, $\lvert S \rvert = 50\,000$.
For observables depending on the visible-unit indices $i, i_1, i_2$,
the data are averaged over all those indices.
Furthermore, as in the main text, each data point is an average over five independently trained RBMs at a fixed training epoch $t$.
The gray dashed line shows the function $\tau_\theta^{(f)} = \tau_\theta$ for reference.}
\label{fig:S:PN:ObsIACTCompare}
\end{figure*}

\section{Autocorrelation times of different observables}
\label{ssec:ObsDepOfIACT}

As explained in the main text (see Methods),
different observables $f(x)$ generally exhibit different integrated autocorrelation times $\tau_\theta^{(f)}$.
We recall that the latter are defined as
\begin{equation}
\label{eq:S:iact:General}
	\tau_\theta^{(f)} := 1 + 2 \sum_{n=1}^\infty \frac{g^{(f)}_\theta(n)}{g^{(f)}_\theta(0)} \,,
\end{equation}
where
\begin{equation}
\label{eq:S:cf:General}
	g_\theta^{(f)}(n) := \< f(x^{(0)}) f(x^{(n)}) \> - \< f(x^{(0)}) \>^2
\end{equation}
is the correlation function of $f(x)$ for the Markov chain~\eqref{eq:S:MarkovChainXH}
initialized in the stationary state $x^{(0)} \sim \hat{p}_\theta(x)$
(cf.\ Eqs.~\meqref{eq:cf:General} and~\meqref{eq:iact:General} in the main text).

The quantity $\tau_\theta$ from Eq.~\meqref{eq:iact},
which is our principal measure of sampling efficiency in the main text,
is a weighted average of the autocorrelation times $\tau_\theta^{(x_i)}$ of the individual visible units $f(x) = x_i$ (see Eq.~\meqref{eq:iactAsVisUnitAvg}).
In the following, we verify that the autocorrelation times $\tau_\theta^{(f)}$ typically scale similarly to $\tau_\theta$.
More precisely, they are found to be largely proportional to each other.
For the power-law bound~\meqref{eq:Tradeoff} which quantifies the accuracy--efficiency tradeoff,
this changes the constant $c$ on the right-hand side to an observable-dependent $c^{(f)}$.
Those constants can no longer be identified with the total correlation $\Ctot(p)$ of the full target distribution $p(x)$.
Instead, the pertinent reference should be a characteristic of the distribution $p(f(x))$ of the transformed variables,
for which, however, it may not always be possible or reasonable to define a ``total correlation.''

In Fig.~\ref{fig:S:TFIC:ZBasis:ObsIACTCompare},
we adopt the same setup and RBMs as in the third column of Fig.~\mref{fig:Tradeoff:TFIM}b of the main text (TFIC, $\sigma^z$ basis, $\nCD = 1$, $\lvert S \rvert = 25\,000$).
We plot the integrated autocorrelation times $\tau_\theta^{(f)}$ for various observables against $\tau_\theta$.
Concretely, the investigated observables are
\begin{subequations}
\label{eq:S:cf:2pointNN}
\begin{itemize}
	\item the individual visible units,
\begin{equation}
	f(x) = x_i \,;
\end{equation}	
	\item the nearest-neighbor correlation function,
\begin{equation}
	f(x) = x_i x_{i+1} \,;
\end{equation}
	\item the next-nearest neighbor correlation function,
\begin{equation}
	f(x) = x_i x_{i+2} \,;
\end{equation}
	\item the $3$-point nearest-neighbor correlation function,
\begin{equation}
	f(x) = x_i x_{i+1} x_{i+2} \,.
\end{equation}
\end{itemize}
\end{subequations}
Note that, due to translational invariance of the target distribution $p(x)$,
all those observables should be independent of the reference index $i \in \{ 1, \ldots, \NV \}$.
However, the learned model distribution $\hat p_\theta(x)$ may not fully reflect this symmetry.
The autocorrelation time $\tau_\theta^{(f)}$ shown in Fig.~\ref{fig:S:TFIC:ZBasis:ObsIACTCompare} is therefore averaged over all indices $i$.
We also show the autocorrelation time for the general $2$-point correlation function
\begin{equation}
	f(x) = x_{i_1} x_{i_2} \,, \quad i_1, i_2 \in \{ 1, \ldots \NV \} \,,
\end{equation}
again averaged over all pairs $(i_1, i_2)$.
(Note that this observable still depends on the difference $i_1 - i_2$ though.)
Finally, we also include autocorrelation times for
the mean over all visible units,
\begin{equation}
\label{eq:S:cf:mean}
	f(x) = \frac{1}{M} \sum_i x_i \,.
\end{equation}
We point out that there are RBM configurations for which we did not obtain a reliable estimate of $\tau_\theta$ or $\tau_\theta^{(f)}$ within the maximally admitted number of sampling steps (see also Methods in the main text).
Therefore, the data points are sparser for $\NH = 64$, in particular.

As mentioned above,
the results indicate that $\tau_\theta^{(f)}$ is usually proportional to $f(x)$.
With regard to the seemingly largest (relative) deviations in the top-left panel,
we observe that the autocorrelation times are generally very small in this case.
We also remark that, due to translation invariance,
the trained models should satisfy $\tau_\theta^{(x_i)} = \tau_\theta$ if they realized this symmetry exactly.
Deviations from this ideal behavior can hint at overfitting and insufficient expressivity.

The same conclusions can be drawn from Fig.~\ref{fig:S:PN:ObsIACTCompare},
which shows autocorrelation times for different observables in the digit-pattern example from Fig.~\mref{fig:Tradeoff:Applications}a--c and Sec.~\ref{ssec:Examples:PN}.
We remark that, due to the distinct geometry, some of the observables do not have the same physical meaning as in the TFIC example;
for instance, $x_i x_{i+1}$ is a correlation function between nearest neighbors along the columns only.

\begin{figure*}
\centering
\includegraphics[scale=1]{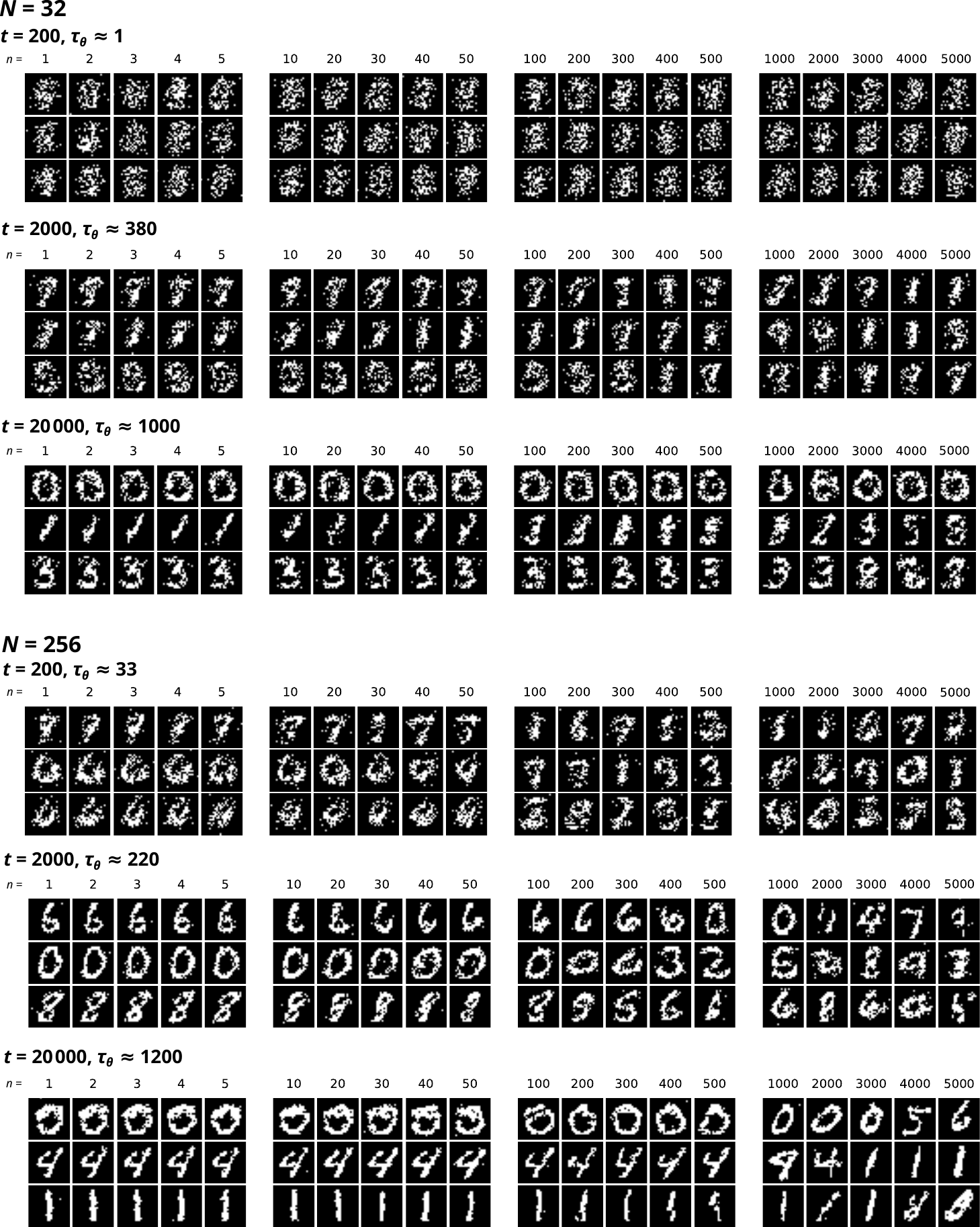}
\caption{Example Markov chains obtained from RBMs trained on the MNIST dataset with the PCD algorithm and $\nCD = 1$, $\eta = 10^{-4}$, $B = 100$ for different numbers of hidden units $\NH$ and training epochs $t$.
All chains were initialized with a uniform random distribution of the visible units $x_i$ [i.e., $p(x_i = 0) = p(x_i = 1) = \frac{1}{2}$]
and subsequently thermalized for $2\times 10^6$ steps before recording starts.}
\label{fig:S:MNIST:MCChains}
\end{figure*}

Finally, for a more direct visualization of correlations between samples of Markov chains like~\eqref{eq:S:MarkovChainXH},
we show snapshots from such chains for RBMs trained on the MNIST dataset in Fig.~\ref{fig:S:MNIST:MCChains}.
The autocorrelation-time estimate $\tau_\theta$ conforms nicely with the number of steps needed to reach a sample that looks ``new'' or ``uncorrelated'' to the naked eye.
The adequacy of $\tau_\theta$ to quantify correlations between Markov-chain samples and to estimate the additional steps required to obtain an independent sample is thus reinforced.

\section{Extended mechanism}
\label{ssec:XMechanism}

\begin{figure}
\centering
\includegraphics[scale=1]{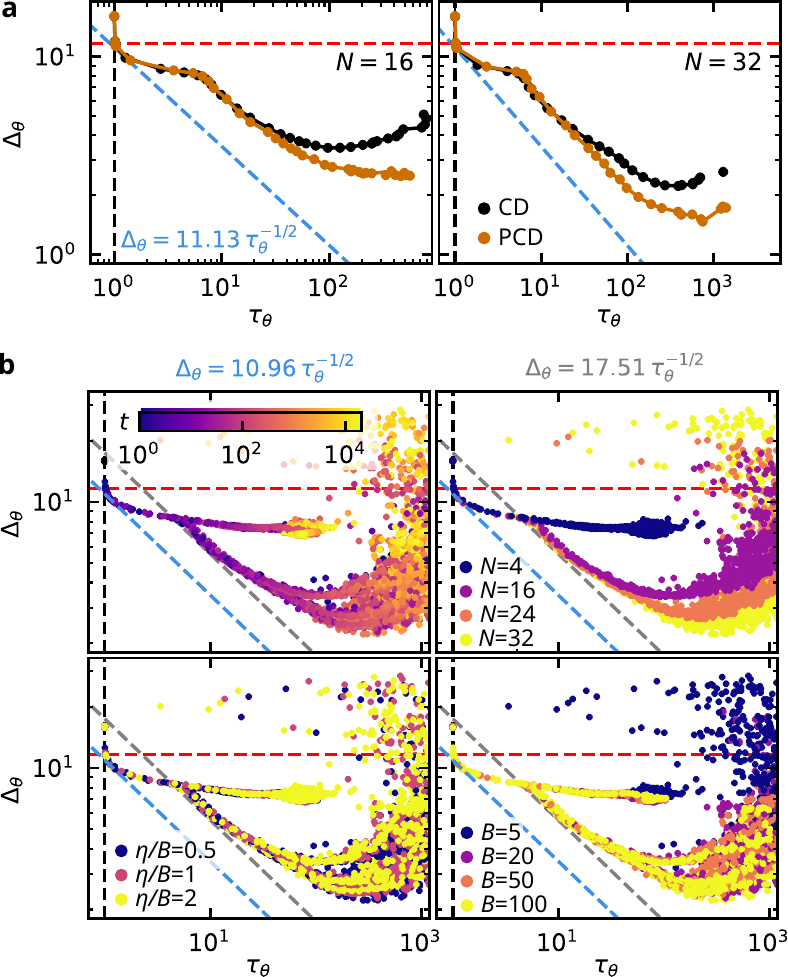}
\caption{Exact loss $\Delta_\theta$ vs.\ autocorrelation time $\tau_\theta$ for RBMs in the digit-generator example from Fig.~\mref{fig:Tradeoff:Applications}a--c and Sec.~\ref{ssec:Examples:PN}.
Training used a dataset of $\lvert S \rvert = 50\,000$ samples.
Data points are averages over $5$ independent runs.
(a) Comparison of ordinary contrastive divergence (CD) and persistent contrastive divergence (PCD) with $\nCD = 1$, $\eta = 0.005$, $B = 100$ for $\NH = 16$ (left) and $\NH = 32$ (right).
(b) CD training with $\nCD = 1$ and $\NH = 4, 16, 24, 32$, $\eta/B = 0.5, 1, 2$, $B = 5, 20, 50, 100$.
All panels show the same data points, but highlight different hyperparameter dependencies by color and brightness as indicated.
Overlapping curves of different colors thus signal that the learning characteristics are indepenent of the respective hyperparameter.}
\label{fig:S:Tradeoff:PN}
\end{figure}

We expand on aspects of the discussion in the section ``Mechanism behind the learning stages'' from the main text.

\subsection{Minimal loss for independent units}
\label{ssec:XMechanism:MinLossIndep}

As stated around Eq.~\meqref{eq:Ctot} of the main text,
the exact loss $\Delta_\theta = \DKL(p || \hat p_\theta)$ is bounded from below by the total correlation $\Ctot(p)$ if $\hat p_\theta$ consists of independent units.
Indeed, if $\hat p(x) := \prod_i \hat p_i(x_i)$,
we can make the following decomposition:
\begin{align}
	\DKL(p || \hat p) 
	&= \sum_x p(x) \left[ \log \frac{p(x)}{\prod_i p_i(x_i)} + \log \frac{\prod_i p_i(x_i)}{\prod_i \hat p_i(x_i)} \right] \notag \displaybreak[0] \\
	&= \sum_x p(x) \log \frac{p(x)}{\prod_i p_i(x_i)}
		+ \sum_i \sum_{x_i} p_i(x_i) \log \frac{ p_i(x_i) }{ \hat p_i(x_i) } \notag \displaybreak[0] \\
	&= \Ctot(p) + \sum_i \DKL(p_i || \hat p_i) \,.
\end{align}
Recalling the definition~\meqref{eq:Ctot}
and the fact that $\DKL(p || q) \geq 0$ for arbitrary distributions $p$ and $q$,
we can conclude that $\DKL(p || \hat p) \geq \Ctot(p)$.

\subsection{Hyperparameter independence of the tradeoff relation}

In the main text,
we argued that the relationship between $\Delta_\theta$ and $\tau_\theta$ in the independent- and correlation-learning regimes is essentially independent of the basic RBM hyperparameters,
including the learning rate $\eta$,
the batch size $B$,
the number of training samples $\lvert S \rvert$,
the number of hidden units $\NH$,
and the approximation scheme for model averages during training (CD vs.\ PCD and their order $\nCD$).
Further evidence for this insensitivity is provided in Fig.~\ref{fig:S:Tradeoff:PN}
for the digit-generator example from Fig.~\mref{fig:Tradeoff:Applications}a--c and Sec.~\ref{ssec:Examples:PN}.

We emphasize that the choice of appropriate hyperparameters is still important,
because it affects the stability of training
and the onset of the degradation regime,
meaning that poor choices can lead to early deterioration of the RBMs.

\subsection{Scaling of weights and correlations}
\label{ssec:XMechanism:ScalingOfWeights}

\begin{figure*}
\centering
\includegraphics[scale=1]{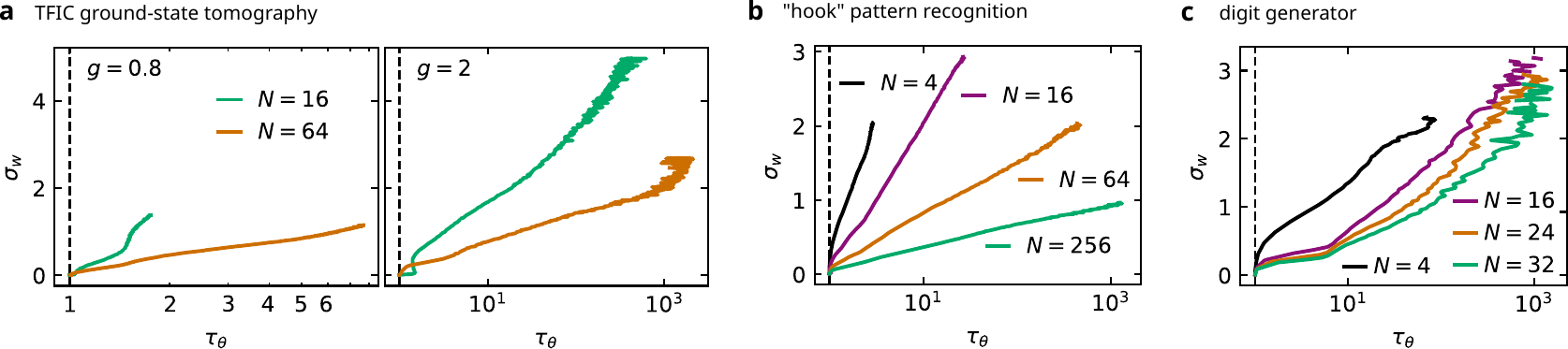}
\caption{Standard deviation $\sigma_w := (\frac{1}{\NV\NH - 1} \sum_{i,j} w_{ij}^{\,2})^{1/2}$ of the weights $w_{ij}$ vs.\ integrated autocorrelation time $\tau_\theta$ for RBMs trained with contrastive divergence of order $\nCD = 1$ on three different problems.
(a) Ground-state tomography in the transverse-field Ising chain of $M = 20$ sites (cf.\ Fig.~\mref{fig:Tradeoff:TFIM} and Sec.~\ref{ssec:Examples:TFIM}),
$\lvert S \rvert = 25\,000$, $\eta = 10^{-3}$, $B = 100$, different $\NH$ and $g$ as indicated.
(b) Pattern-recognition task (cf.\ Fig.~\mref{fig:Tradeoff:PH1} and Sec.~\ref{ssec:Examples:PH1}),
$\lvert S \rvert = 5000$, $\eta = 0.005$, $B = 100$, different $\NH$ as indicated.
(c) Digit generator (cf.\ Fig.~\mref{fig:Tradeoff:Applications}a--c and Sec.~\ref{ssec:Examples:PN}),
$\lvert S \rvert = 50\,000$, $\eta = 0.005$, $B = 100$, different $\NH$ as indicated.}
\label{fig:S:WSTDvsACT}
\end{figure*}

The magnitude $\lvert \theta_k \rvert$ of the RBM parameters typically grows during training.
A distinctive property of many ``real-world'' machine-learning problems is that the target distribution is sparse,
meaning that most states $x$ have vanishing or at least very small probability $p(x)$.
For example, the overwhelming majority of all possible images with a given number of pixels will not display ``realistic'' motifs (e.g., digits, letters, animals, clothes, buildings, ...).
This characteristic is at odds with the RBM model family,
which assigns a finite probability $\hat p_\theta(x) > 0$ to all states $x$.
To suppress the unlikely states,
many of the parameters $\theta_k$ have to take large absolute values.

As argued around Eq.~\meqref{eq:pHGivenXEntropy}  and illustrated in Fig.~\ref{fig:S:WSTDvsACT},
this usually increases the correlations between subsequent Markov-chain samples.
Notably,
correlations between visible and hidden units,
and thus between two (or more) visible units,
arise only if $\lvert w_{ij} \rvert > 0$.
Hence larger $\lvert \theta_k \rvert$ and larger $\lvert w_{ij} \rvert$ in particular 
hint at larger autocorrelation times $\tau_\theta$.
Note, however, that $\lvert w_{ij} \rvert > 0$ for some $i, j$ is not sufficient to obtain correlations between different visible units;
to this end, two visible units $x_{i_1}$ and $x_{i_2}$ must be coupled to the same hidden unit $h_j$, i.e., both $\lvert w_{i_1 j} \rvert > 0$ and $\lvert w_{i_2 j} \rvert > 0$ must hold.

Since it is computationally demanding to estimate $\tau_\theta$ reliably,
the standard deviation $\sigma_w := (\frac{1}{\NV\NH - 1} \sum_{i,j} w_{ij}^{\,2})^{1/2}$
can be considered as a more accessible indicator of growing correlations in practice.
In Fig.~\ref{fig:S:WSTDvsACT}, we show the mutual dependence of $\tau_\theta$ and $\sigma_w$ for the examples from Sec.~\ref{ssec:Examples:TFIM}--\ref{ssec:Examples:PN} and various numbers of hidden units $\NH$.
We observe that the two quantities are indeed positively correlated,
which confirms, in particular, that $\tau_\theta$ typically grows with the magnitude of the weights.

The necessity to increase the RBM parameters $\theta_k$ in magnitude in order to represent sparse distributions with many ``inactive'' states $x$ such that $p(x) = 0$
is also one possible hint at the source of the different $\alpha$ values observed in Fig.~\mref{fig:Tradeoff:TFIM}d compared with Fig.~\mref{fig:Tradeoff:TFIM}b (and also Figs.~\mref{fig:Tradeoff:PH1}b and~\mref{fig:Tradeoff:Applications}b,c).
The image distributions from Figs.~\mref{fig:Tradeoff:PH1}a--c and~\mref{fig:Tradeoff:Applications}b,c all have a large fraction of such ``inactive'' states ($99.92\,\%$ and $99.88\,\%$, respectively, see Secs.~\ref{ssec:Examples:PH1} and~\ref{ssec:Examples:PN}).
Likewise, the TFIC ground state in the $\sigma^z$ basis (when $\alpha = \frac{1}{2}$) has $p(x) = 0$ for half of the states by symmetry (see Sec.~\ref{ssec:Examples:TFIM} and~\ref{ssec:TFICXBasis}),
whereas no such inactive states exist in the $\sigma^x$ representation (when $\alpha = 6 \ldots 8$).

Within the examples from Fig.~\mref{fig:Tradeoff:TFIM}d,
the case with $g = 1$ stands out
because it also exhibits a pronounced intermediate stage in the correlation-learning regime
where the relationship between $\Delta_\theta$ and $\tau_\theta$ does not follow the power-law tradeoff of the global bound with $\alpha \simeq 6$,
but instead shows a stronger tradeoff with $\alpha \simeq \frac{1}{4}$.
A more detailed analysis (see Sec.~\ref{ssec:TFICXBasis}) suggests that this is caused by the emerging strong bimodal structure of the target distribution as $g$ becomes smaller,
which impedes and eventually prevents efficient learning.

\subsection{Basis-dependent learning characteristics in the transverse-field Ising chain}
\label{ssec:TFICXBasis}

\begin{figure*}
\centering
\includegraphics[scale=1]{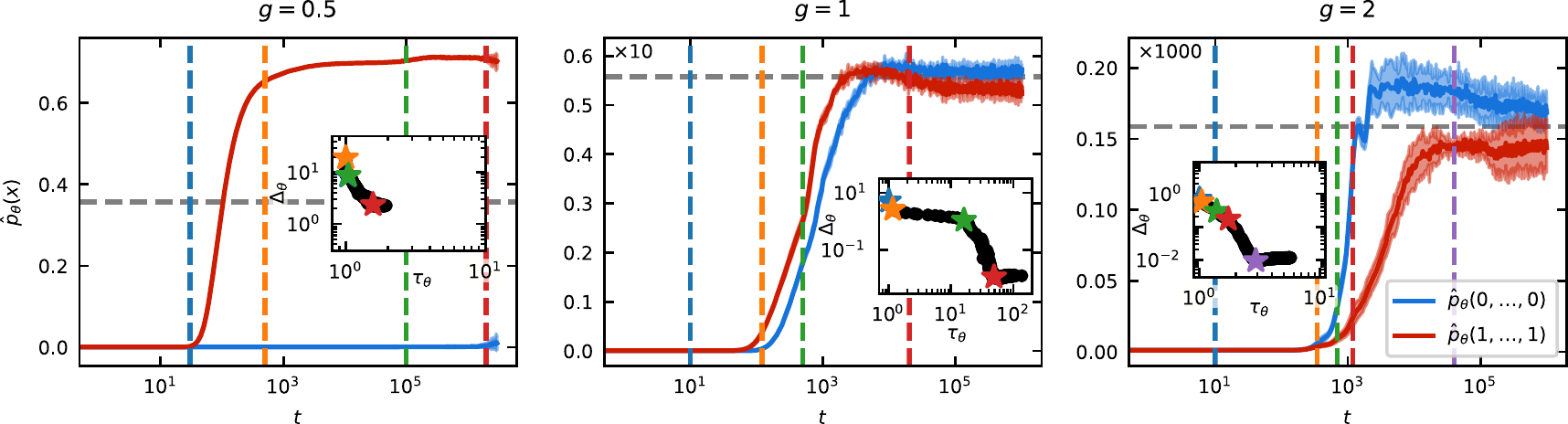}
\caption{Evolution of the model probabilities $\hat p_\theta(x)$ with the training time $t$ for the two dominant states of the target distribution $p^x(x)$,
$x = (0, \ldots, 0)$ (all spins up, blue) and $x = (1, \ldots, 1)$ (all spins down, red).
Horizontal dashed lines show $p^x(x)$ for those states.
Shaded areas indicate the fluctuations across five independent runs (standard deviation).
Training used contrastive divergence of order $\nCD = 1$ and hyperparameters as in Fig.~\mref{fig:Tradeoff:TFIM}d of the main paper, i.e., $\eta = 10^{-3}$, $B = 100$, $\lvert S \rvert = 25\,000$.
Insets: Exact loss $\Delta_\theta$ vs.\ autocorrelation time $\tau_\theta$.
Selected epochs are marked by colored vertical dashed lines in the main panel and stars in the insets.
Note the scaling of the $y$-axis in the middle and right panels as indicated in the top-left corner.
}
\label{fig:S:TFIC:XBasis:DomStateProbs}
\end{figure*}

We investigate the emergence of an intermediate stage in the correlation-learning regime,
where learning is less efficient than admitted by the global power-law tradeoff,
in the example of ground-state tomography for the transverse-field Ising chain in the $\sigma^x$ basis
as $g$ becomes smaller
(see, in particular, Fig.~\mref{fig:Tradeoff:TFIM}d of the main text).

Since there are some important differences in the learning characteristics of the TFIC ground-state distribution in the $\sigma^z$ and $\sigma^x$ bases (cf.\ Figs.~\mref{fig:Tradeoff:TFIM}b and~d),
we first examine the underlying target distribution $p(x)$ in more detail.
To avoid confusion in the following,
we denote the target distribution as $p^z(x)$ when referring to the $\sigma^z$ basis representation
and as $p^x(x)$ when referring to the $\sigma^x$ basis representation.
The Hamiltonian~\eqref{eq:S:TFIM:H} has a $\ZZ_2$ symmetry of the form $F := \prod_i \sigma_i^z$,
i.e., $[H, F] = 0$.
In the $\sigma^z$ basis,
$F$ measures the parity of the number of down spins,
whereas it describes a spin-flip symmetry in the $\sigma^x$ basis.

For even $M$,
to which we restrict our discussion exclusively in this work,
the ground state~\eqref{eq:S:TFIM:GroundState} of the Hamiltonian~\eqref{eq:S:TFIM:H} lies in the $F = +1$ sector.
In the $\sigma^z$ basis,
this means that $p^z(x) = 0$ for half of the basis states,
namely all configurations $x = (x_1, \ldots, x_{\NV})$ for which $\sum_i x_i$ is odd.
In the $\sigma^x$ basis, by contrast, it implies that $p^x(x) = p^x(1 - x)$,
where $1 - x = (1 - x_1, \ldots, 1 - x_{\NV})$.
This distinct manifestation of the symmetry $F$ is the first important distinction between the $\sigma^z$- and $\sigma^x$-basis representations.
Learning the $\sigma^x$ representation is thus aided by the fact that
the RBM model can encode \emph{relative} spin orientations (alignment or anti-alignment) particularly efficiently,
requiring only a single hidden unit for an arbitrary combination of visible ones,
although the effectively needed number of weights appears to depend also on the proximity to the critical point \cite{Golubeva:2021prb}.
By contrast, due to the exponential form of the model distribution, it is rather inefficient at enforcing $\hat p_\theta(x) = 0$ for individual states $x$ as required by the $\sigma^z$ representation $p^z(x)$.
These observations hint at possible origins of the different exponents $\alpha$ in the tradeoff relation~\meqref{eq:Tradeoff} for the two representations,
a large value $\alpha \simeq 6 \ldots 8$ for the better suited $p^x(x)$
and a small value $\alpha \simeq \frac{1}{2}$ for the worse suited $p^z(x)$.

For large values of $g$,
$p^z(x)$ is dominated by the state $x = (0, \ldots, 0)$ (``all spins up''),
while $p^x(x)$ becomes approximately uniform.
For small values of $g$, in turn,
$p^z(x)$ approaches a uniform distribution,
whereas $p^x(x)$ is dominated by the two states $x = (0, \ldots, 0)$ and $x = (1, \ldots, 1)$,
whose probability is degenerate due to the symmetry $F$.
The different mode structure in the nonuniform limit of the $\sigma^z$- and $\sigma^x$-basis representations is a second important difference between them.
Strongly unimodal as well as approximately uniform distributions are structurally simple,
and gradient-descent training can find a globally optimal solution relatively easily.
If the distribution has two (or a few) modes or dominant states,
by contrast,
the risk of getting stuck in a local minimum of the loss landscape and detecting only a subset of those dominant states is increased.

We presume that the bimodal structure of $p^x(x)$,
which becomes increasingly pronounced as $g$ becomes smaller,
is the reason for the intermediate stage in the correlation-learning regime observed in the top panel ($g = 1$) of Fig.~\mref{fig:Tradeoff:TFIM}d.
To substantiate this claim,
we investigate the evolution of the model probability $\hat p_\theta(x)$ during training for the two modes $x^+ := (0, \ldots, 0)$ and $x^- := (1, \ldots, 1)$.
Fig.~\ref{fig:S:TFIC:XBasis:DomStateProbs} shows these probabilities for three different values of $g$.

We focus on the middle panel ($g = 1$) first,
where $p^x(x^\pm) \approx 5.6\,\%$.
The end of the correlation-learning regime is marked by the blue dashed line (blue star in the inset).
Thereafter,
the RBM first picks up the $x^-$ mode,
accompanied by a segment with $\alpha \simeq 6$
(see also Fig.~\mref{fig:Tradeoff:TFIM}d in the main text).
The probability of the other mode $x^+$ only starts to increase slightly later at around the orange mark,
which also corresponds to the starting point of the second stage of correlation learning with $\alpha \simeq \frac{1}{4}$.
It lasts until both modes are established (green mark),
and the subsequent fine tuning proceeds with a more efficient tradeoff exponent of $\alpha \simeq 6$ again. 
A similar, but less pronounced behavior can be observed for $g = 2$ (right panel of Fig.~\ref{fig:S:TFIC:XBasis:DomStateProbs}),
where the two modes only have $p^x(x^\pm) \approx 0.015\,\%$.
On the contrary, if the modes become more pronounced as exemplified by the $g = \frac{1}{2}$ case with $p^x(x^\pm) \approx 36\,\%$ (left panel),
the RBM only learns one of the modes and fails to detect the other one.

\end{document}